\documentclass[journal]{IEEEtran}
\ifCLASSINFOpdf
\else
\fi

\usepackage{cite}
\usepackage{xspace}
\usepackage{tabularx}
\usepackage{algorithm}
\usepackage{algorithmic}
\usepackage{epsfig}
\usepackage{graphicx}
\usepackage{amsmath}
\usepackage{amssymb}
\usepackage{mathtools}
\usepackage{subfigure}
\usepackage{multirow}
\usepackage{makecell}
\usepackage{color}
\usepackage{float}
\usepackage{picinpar}
\usepackage{bbding}
\usepackage{url}

\usepackage[misc]{ifsym}
\usepackage{tablefootnote}
\usepackage{booktabs}

\usepackage{siunitx}
\newcommand{\etal}{\textit{et al}.}
\newcommand{\ie}{\textit{i}.\textit{e}.}
\newcommand{\eg}{\textit{e}.\textit{g}.}

\newcommand{\slfrac}[2]{\left.#1\middle/#2\right.}

\begin{document}
%
\title{Automatic Sparse Connectivity Learning for Neural Networks}
%
%
%

\author{Zhimin~Tang, Linkai~Luo$^{*}$, Bike~Xie, Yiyu~Zhu, Rujie~Zhao, Lvqing~Bi, Chao~Lu$^{*}$
\thanks{Z. Tang and L. Luo are with Department of Automation, Xiamen University, Xiamen, 361102, China. Tang is a visiting scholar at the ECE Department of SIUC from 2018 to 2020. (e-mail: luolk@xmu.edu.cn)}
\thanks{B. Xie and Y. Zhu are with Kneron Inc. San Diego, CA, 92121, USA.}
\thanks{L. Bi is with School of Physics and Telecommunication Engineering, Research Center for Intelligent Information and Communication Technology, Yulin Normal University, Yulin 537000, Guangxi, China.}
\thanks{R. Zhao and C. Lu are with the Department of Electrical and Computer Engineering, Southern Illinois University Carbondale (SIUC), Carbondale, IL, 62091, USA. (e-mail: chaolu@siu.edu)}
\thanks{$^{*}$Corresponding authors: L. Luo and C. Lu}
}

%
%

\markboth{IEEE Transactions on Neural Networks and Learning Systems}%
{Tang \MakeLowercase{\textit{et al.}}: Automatic Sparse Connectivity Learning for Neural Networks}
%


\maketitle

\begin{abstract}
Since sparse neural networks usually contain many zero weights, these unnecessary network connections can potentially be eliminated without degrading network performance. Therefore, well-designed sparse neural networks have the potential to significantly reduce FLOPs and computational resources. In this work, we propose a new automatic pruning method - Sparse Connectivity Learning (SCL). Specifically, a weight is re-parameterized as an element-wise multiplication of a trainable weight variable and a binary mask. Thus, network connectivity is fully described by the binary mask, which is modulated by a unit step function. We theoretically prove the fundamental principle of using a straight-through estimator (STE) for network pruning. This principle is that the proxy gradients of STE should be positive, ensuring that mask variables converge at their minima. After finding Leaky ReLU, Softplus, and Identity STEs can satisfy this principle, we propose to adopt Identity STE in SCL for discrete mask relaxation. We find that mask gradients of different features are very unbalanced, hence, we propose to normalize mask gradients of each feature to optimize mask variable training. In order to automatically train sparse masks, we include the total number of network connections as a regularization term in our objective function. As SCL does not require pruning criteria or hyper-parameters defined by designers for network layers, the network is explored in a larger hypothesis space to achieve optimized sparse connectivity for the best performance. SCL overcomes the limitations of existing automatic pruning methods. Experimental results demonstrate that SCL can automatically learn and select important network connections for various baseline network structures. Deep learning models trained by SCL outperform the state-of-the-art human-designed and automatic pruning methods in sparsity, accuracy, and FLOPs reduction.
\end{abstract}


%
\IEEEpeerreviewmaketitle

\section{Introduction}
%
%
%
%
\IEEEPARstart{D}{ESPITE} the great success in improving neural networks and learning systems \cite{szegedy2015going,krizhevsky2012imagenet,simonyan2014very,he2016deep,huang2017densely,lecun2015deep}, state-of-the-art deep neural networks usually consist of dozens of stacked layers and a huge number of parameters. It is difficult to deploy these over-parameterized and over-redundant neural networks on resource-constrained computing platforms \cite{han2015learning}. To address this challenge, network pruning has received great attention. Network pruning tends to remove unimportant trainable parameters in a neural network architecture while maintaining its high accuracy. Effective network pruning leads to less computing operations, memory usage, and power consumption with little performance degeneration. After performing network pruning, sparse models are often implemented in hardware (\eg, GPUs, ASICs\footnote{\url{https://www.kneron.com/solutions/soc/}} or FPGAs\footnote{\url{https://www.xilinx.com/applications/megatrends/machine-learning.html}}) for AI acceleration. 

Network pruning can be divided into two categories: human-designed pruning and automatic pruning. The former requires some pruning criteria or hyper-parameters defined by designers (\eg, importance measure or pruning threshold), while automatic pruning generates optimized sparse networks with little human intervention. Human-designed pruning usually consists of three steps: (1) training weight parameters in a selected baseline neural network, (2) eliminating unimportant network connections based on designer-defined criteria or hyper-parameters, and (3) training weight parameters again in this pruned network architecture. Human-designed network pruning can be further divided into two types: unstructured  \cite{han2015learning,guo2016dynamic,ullrich2017soft,molchanov2017variational,zhu2017prune,tartaglione2018learning,frankle2018lottery} or structured \cite{li2016pruning,luo2017thinet,he2018soft,lin2018accelerating,lin2019toward,he2019filter,liu2017learning,he2017channel,zhuang2018discrimination,liu2019channel,li2019oicsr,wen2016learning,anwar2017structured}. It has been reported that performing unstructured pruning on deep neural networks does not cause much loss of accuracy. On the other hand, structured pruning especially filter \cite{li2016pruning,luo2017thinet,he2018soft,lin2018accelerating,lin2019toward,he2019filter} and channel \cite{liu2017learning,he2017channel,zhuang2018discrimination,liu2019channel,li2019oicsr} pruning, has been used to accelerate neural networks in general hardware platforms (\ie, GPUs). Since zero-masked feature maps can be deleted, less computational cost is required after network pruning. Note that some existing neural networks with multi-branch or multi-group structures may be considered as human-designed sparse architectures (\eg, Inception \cite{szegedy2015going}, ShuffleNet \cite{zhang2018shufflenet}, MobileNet \cite{howard2017mobilenets}, ResNeXt \cite{xie2017aggregated}), even though these sparse architectures do not involve network pruning. 

Up to date, three major bottlenecks have hindered the use of human-designed pruning methods to generate sparse networks. First, although human-designed pruning can provide good network performance under a low pruning rate, the network performance under a high pruning rate is severely degraded. Second, there is a lack of appropriate methods to effectively prune network connections for high-compression and high-performance neural networks. In most human-designed pruning methods, network connections are pruned based on the assumption of “smaller-norm-less-important”. Network connections with smaller weights are generally considered trivial and eliminated during network pruning. However, researchers have found that sometimes the amount of information in smaller weights is important and cannot be ignored \cite{ye2018rethinking}. Therefore, considering the variability of loss function sensitivity with respect to different weights, eliminating network connections with smaller weights does not guarantee a slight decrease in network performance. Third, the biggest weakness is the need for designer-defined pruning criteria (\eg, $L_{1}$ \cite{han2015learning} or $L_{2}$ \cite{he2018soft} norm of filters) and hyper-parameters (\eg, pruning threshold or ratio) for each network layer during network pruning. Because the selection of pruning criteria or hyper-parameters is manually determined, it heavily depends on the designer's prior experience and varies from application to application. Consequently, since the pruning criteria and hyper-parameters are not guaranteed to be the best choice, the resulting network connectivity and performance are not optimal. Note that even though weight importance estimation methods have been proposed in \cite{molchanov2017pruning,theis2018faster}, layer-wise hyper-parameters are still needed to determine a pruning threshold or ratio for all layers during network pruning. Unfortunately, as the optimal pruning threshold or ratio varies with local network structures, the use of only one hyper-parameter for all layers leads to inferior pruning performance. Recently, it is proposed that a proper criterion may be selected from a set of designer-defined criteria to learn pruning criteria for each network layer \cite{he2020learning}. Yet, the pruning performance of this method still depends on the quality of candidate criteria defined by designers.

To get rid of shortcomings of human-designed pruning, researchers have investigated automatic pruning, which trains sparse network connectivity through task-aware loss or a sparse regularization term \cite{louizos2017learning,kang2020operation,herrmann2020channel,huang2018data,xiao2019autoprune}. Note automatic pruning does not require designer-defined pruning criteria or layer-wise hyper-parameters. 
Louizos \etal \cite{louizos2017learning} train sparse neural networks through $L_{0}$-norm regularization. They use the Gumbel-Softmax trick \cite{jang2017categorical,maddison2017concrete} (also known as a concrete distribution) and apply gates to weights for connectivity training. Note a stochastic gate produces zero connectivity only when the probability is zero, which is almost impossible to reach during network training. To address this problem, threshold operation is applied on gates to made zero connectivity. The drawback of \cite{louizos2017learning} is that the expected $L_{0}$-norm of stochastic training does not reflect the $L_{0}$-norm of deterministic inference. Therefore, as will be discussed in Section \ref{sec:l0}, although the expected $L_{0}$ norm of weights is significantly reduced, it is still not low enough to produce sparse connections. Kang \etal \cite{kang2020operation} propose soft channel pruning (SCP), which assumes that feature maps follow a Gaussian distribution and features are pruned if the cumulative density function is larger than a certain threshold. The Gumbel-Softmax trick is used to tackle the non-differentiable Bernoulli distribution sampling. Unfortunately, the assumption of Gaussian distribution for feature maps is too strict to derive accurate gradients. Moreover, SCP shows good pruning performance in network layers that are followed by both batch normalization (BN) and ReLU. According to evaluation results in \cite{kang2020operation}, its pruning performance is severely degraded if only BN exists. Since many neural networks do not include both BN and ReLU, the application scope of SCP is limited. Herrmann \etal \cite{herrmann2020channel} jointly consider conditional computation and network pruning. The Gumbel-Softmax trick is used to relax the discrete masks to a continuous form. The researchers focus on dynamic inference using conditional computation. Depending on network inputs, masks are dynamically applied on channels. Unfortunately, these masks are data-dependent rather than being fixed, and the dynamically pruned structures are not friendly to hardware implementation. Huang \etal \cite{huang2018data} introduce a series of non-negative scaling factors that are associated with neural network connectivity. To encourage sparse network connectivity, these scaling factors are penalized by an $L_1$-norm regularization term. During the training process, stochastic gradient descent (SGD) and a proposed stochastic Accelerated Proximal Gradient (APG) are used to train weight parameters and scaling factor parameters, respectively. The scaling factor parameters are converted into scaling factors through a soft-threshold operation. Xiao \etal \cite{xiao2019autoprune} utilize STE to relax binary masks. Even though empirical experiments have shown potential, the fundamental principle of using STE in network pruning has not been explored.

From the above discussion, it is clear that existing network pruning methods, either human-designed or automatic, do not fully address the requirement of highly effective spare connectivity learning. It is attractive to develop new pruning methods to overcome the drawbacks and limitations of existing pruning methods. In this work, we propose a new automatic network pruning technique - sparse connectivity learning (SCL).  This work makes the following contributions:
\begin{itemize}
   \item We theoretically prove the fundamental principle of using STE for network pruning. After finding Leaky ReLU, Softplus, and Identity STEs can satisfy this principle, we propose to adopt Identity STE in SCL for discrete mask relaxation. Thus, SCL guarantees the convergence of mask variables at their minima. 
  \item We observe that mask gradients on different features have a wide range of magnitudes and hence are unbalanced. Therefore, we propose to normalize mask gradients of each feature to optimize mask variable training.
  \item The pruning principle of our proposed SCL method is the significance of weight, instead of the magnitude of weight. SCL can automatically learn and determine critical network connections of baseline networks (\eg, DenseNets, ResNets, VGGs, EfficientNets, and RNNs). 
  \item SCL enables highly effective weight-level sparsity learning on neural networks under high pruning rates. Experimental results in the MNIST, CIFAR-10, CIFAR-100, ImageNet, and WikiText-2 datasets demonstrate the enhanced performance of SCL-induced sparse neural networks than the state-of-the-art network pruning methods in the literature.
  \item SCL automatically learns optimized neural network connectivity in a task-aware manner, ensuring that performance-sensitive network connections are ultimately preserved. As it does not need any designer-defined pruning criteria or layer-wise pruning hyper-parameters, SCL gets rid of the limitation of human designers. Compared to the state-of-the-art pruning methods, experimental results demonstrate that SCL results in high-performance neural networks with higher sparsity and fewer FLOPs.
\end{itemize}

Since this work focuses on the algorithm aspect of network pruning, the hardware implementation of spare models generated from our proposed pruning algorithm is beyond the scope of this paper. Sparse neural network models generated from SCL can be implemented in FPGAs or Kneron edge AI hardware products. This paper is organized as follows. Section \ref{sec:related_work} introduces related works about pruning criteria and binary mask relaxation. Section \ref{sec:sparse} describes the proposed automatic sparse connectivity learning theory. Section \ref{sec:baseline} introduces the baseline network architectures and experimental setups. Section \ref{sec:experiments} demonstrates various experimental results and comparisons with the state-of-the-art works in the literature. Section \ref{sec:concusion} concludes the paper. 

\section{Related Work}
\label{sec:related_work}

\subsection{Human-designed pruning criteria}
The performance of human-designed pruning methods depends to a large extent on the quality of pruning criteria defined by designers. Unimportant network connections are removed based on the importance measure, which usually follows the assumption that smaller norms are less important. The criterion for unstructured pruning is the absolute value of weights \cite{han2015learning,guo2016dynamic}. The pruning criterion for structured pruning is the $L1$-norm \cite{li2016pruning} or $L2$-norm \cite{he2018soft} of filter or channel weights. The magnitude of scaling factors in batch normalization is also proposed as the channel pruning criterion \cite{liu2017learning}. He \etal \cite{he2019filter} do not agree that smaller filters are less important, but propose that the contribution of median filters is relatively small, because they can be represented by other filters.

\subsection{Gumbel-Softmax trick}
Automatic pruning involves training binary masks. To address the non-differentiation problem, the Gumbel-Softmax trick~\cite{jang2017categorical,maddison2017concrete} has been used to relax binary masks \cite{louizos2017learning,kang2020operation,herrmann2020channel}. This trick uses gradient methods to train discrete random variables. In this trick, discrete values produced by argmax are encoded in a one-hot vector, and the random sampling $f(U)$ is expressed as
\begin{equation}
\label{eq:gumbel_max}
f(U) = argmax(\log \mathbf{\alpha} + G(U))
\end{equation}
\begin{equation}
\label{eq:gumbel_sampling}
G(U) = -\log(-log U), U \sim \mathcal{U}(0,1)
\end{equation}
where the argmax function finds the argument corresponding to the maximum value, $\alpha$ is a set of unnormalized parameters $\alpha_{k}$ and the probability of outcome $k$ is $\alpha_{k}/{\sum_i \alpha_i}$. Each element in the vector $G(U)$ obeys a Gumbel distribution. As a binary mask has two possible discrete values, mask re-parameterization is regarded as a special case. Due to the sparsity consideration, there is a high possibility that mask sampling results should be trained to be zero. As a result, sparse binary masks are obtained. However, the discrete argmax operation can not be trained by gradient methods. Therefore, one way to circumvent this is to relax the argmax operation by replacing it with softmax \cite{jang2017categorical} and \cite{maddison2017concrete}, as
\begin{equation}
\label{eq:gumbel_softmax}
f(U) = softmax((\log \mathbf{\alpha} + G(U))/\tau)
\end{equation}
where $\tau$ is a hyper-parameter of the softmax function to control relaxation. When $\tau \to 0$, the softmax function becomes the argmax function. Thus, Eq. (\ref{eq:gumbel_softmax}) provides a differentiable form, which is able to train and optimize by gradient methods.

The Gumbel-Softmax trick has two limitations. First, its gradient estimations are biased with respect to the gradients of discrete connectivity. According to Eq. (\ref{eq:gumbel_softmax}), this trick leads to unbiased gradients only when the hyper-parameter $\tau$ tends to 0. However, because the value of $\tau$ is used to balance bias and variance in practice, $\tau$ is rarely chosen to be close to 0. As a result, the studies in \cite{louizos2017learning,kang2020operation,herrmann2020channel} use biased gradients, which may not meet the essential condition for convergence (\ie, low-biased gradient).  
Second, the training process is stochastic, but the inference is deterministic. Therefore, the expected number of connections of stochastic training does not reflect the number of connections of deterministic inference. As shown in Section \ref{sec:l0}, the reduction of $L_0$-norm during training does not mean sparser network connections.

\subsection{Straight-through estimator (STE) trick}

Due to the derivatives of the binarization function are mostly zero, training variables can not update using gradient-based optimization methods. STE is a trick of using proxy gradients in back-propagation \cite{bengio2013estimating}. In this trick, zero gradients of a discrete function are replaced by the derivative of a (sub)differentiable function. Derivatives of ReLU STE and Clipped ReLU STE were used for network quantization \cite{courbariaux2016binarized,yin2019blended,cai2017deep}. Yin \etal \cite{yin2019understanding} referred to proxy gradients as coarse gradients, and studied the STE properties for network quantization. \cite{yin2019understanding} assumes that network inputs obey a normal distribution, yet, this assumption has not been validated for network pruning. Later, based on \cite{yin2019understanding}, Xiao \etal \cite{xiao2019autoprune} proposed to use Leaky ReLU STE and Softplus STE to relax binary masks. Instead of performing network pruning, Hinton \etal \cite{hinton2012neural} proposed to use Identity STE for binary neuron training.
The (sub)differentiable functions of five existing STEs are expressed as
\begin{equation}
\label{eq:relu}
\sigma_{relu}(x) = \max(0, x)
\end{equation}
\begin{equation}
\label{eq:crelu}
\sigma_{clipped\_relu}(x) = \min(\max(0, x), \alpha), \alpha > 0
\end{equation}
\begin{equation}
\label{eq:lrelu}
\sigma_{leaky\_relu}(x) = \max(x, \alpha \cdot x), 0 < \alpha < 1
\end{equation}
\begin{equation}
\label{eq:softplus}
\sigma_{softplus}(x) = \log (1 + \exp(x))
\end{equation}
\begin{equation}
\label{eq:identity}
\sigma_{identity}(x) = x
\end{equation}

The STE trick for network pruning has two limitations. First, as Xiao \etal \cite{xiao2019autoprune} treat connectivity as a hyper-parameter, bi-level optimization is used to update weights and masks separately. Because it involves two-pass calculations to complete an update, the pruning optimization method  is too cumbersome and computationally expensive. Second, although the use of STE for network pruning has achieved empirical success in \cite{xiao2019autoprune}, yet, it is based on the assumption that network inputs obey a normal distribution. Unfortunately, so far, the fundamental principle of using STE for network pruning is still unclear, which is the focus of this work.

\section{Automatic Sparse Connectivity Learning}
\label{sec:sparse}
\begin{figure*}[!t]
\centering
\includegraphics[scale=0.7]{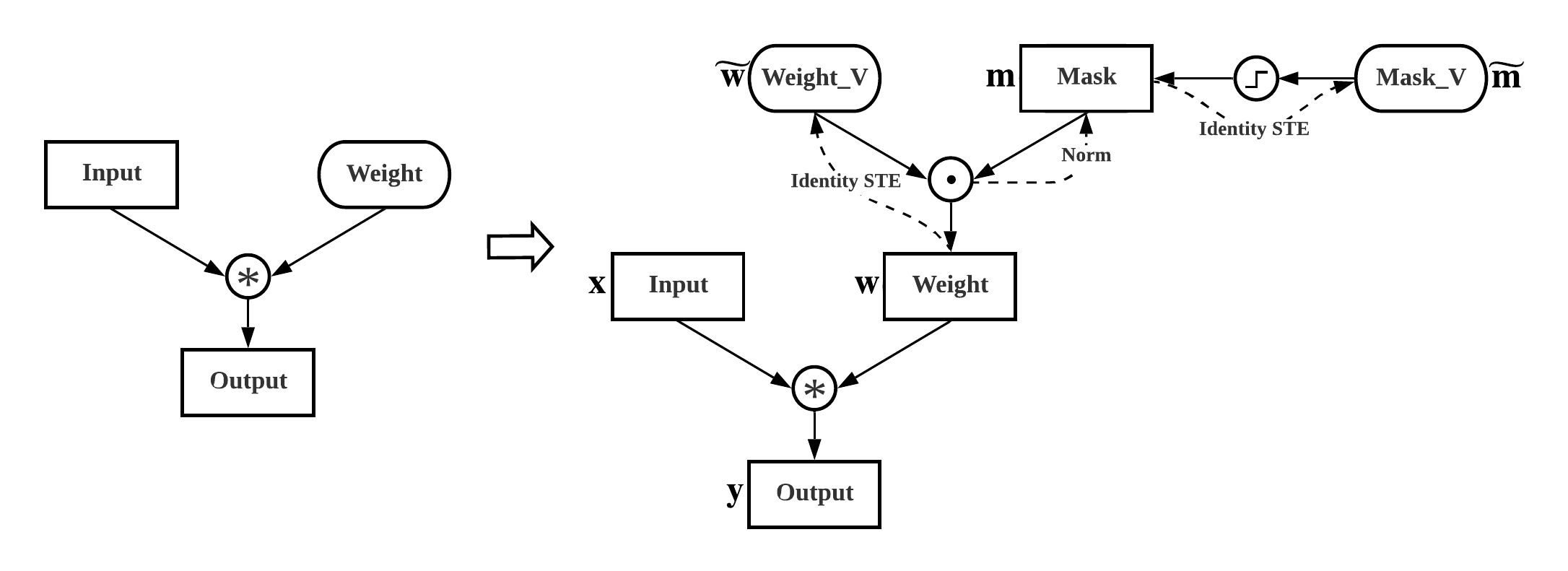}
\caption{Automatic sparse connectivity learning. Weight re-parameterization for sparse convolution, Identity STE for masked weight training, Identity STE for binary mask relaxation, and mask gradient normalization. The elliptical symbols, `Weight\_V' and `Mask\_V', refer to trainable variables, the rectangular symbols indicate intermediate tensors, and the circles indicate calculation operators. Dotted lines mean gradient process during backward propagation.}
\label{fig:reparameterization}
\end{figure*}

The proposed automatic sparse connectivity learning is briefly described below and illustrated in Figure \ref{fig:reparameterization}. First, we represent the weight as a multiplication of a weight variable and a binary mask (0 or 1). 1 indicates a network connection, while 0 indicates no network connection. Thus, the connectivity of a neural network can be fully described by the binary mask, which will be further modulated by a unit step function on a mask variable. Second, during the training process, we will decay the network connectivity to gradually push the elements of the binary mask towards zero (\textit{i.e.}, no network connection). Finally, without the need of designer-defined criteria or layer-wise pruning hyper-parameters, unimportant network connections will be automatically discovered and removed. The design details of each step will be elaborated in the following subsections.

\subsection{Weight re-parameterization}

We design to learn a binary mask that is a standard convolution layer and is related to network connectivity. A fully connected layer is a special case of an input with a feature size of $1 \times 1$ convolved with a kernel with a size of $1 \times 1$. The base cell of a RNN is actually composed of several fully connected modules. We model this convolution operation as
\begin{equation}
\label{eq:conv}
\mathbf{y} = \mathbf{w} * \mathbf{x}
\end{equation}
where the convolutional kernel $\mathbf{w}$ is weight, $\mathbf{x}$ is input, and $\mathbf{y}$ is output. $*$ annotates the convolution operation. As shown in Figure \ref{fig:reparameterization}, a weight is re-parameterized by a weight variable $\mathbf{\widetilde w}$ and a binary mask $\mathbf{m}$. Weight re-parameterization is formulated as
\begin{equation}
\label{eq:mask}
\mathbf{w} = \mathbf{\widetilde w} \odot \mathbf{m}
\end{equation}
where $\odot$ is the element-wise multiplication. Mask is binarized from the trainable mask variable $\mathbf{\widetilde m}$ using a unit step function. The binarization is formulated as
\begin{equation}
\label{eq:step}
\mathbf{m} = H(\mathbf{\widetilde m}) = 
\left\{
\begin{array}{l} 
0, \widetilde m \leq 0 \\
1, \widetilde m > 0
\end{array}
\right.
\end{equation}
Therefore, the elements of $\mathbf{\widetilde w}$ will be zero-masked if the corresponding elements in $\mathbf{\widetilde m}$ is non-positive. 

\subsection{Gradient redefinition for weight variables}

According to the derivative chain rule and Eq. (\ref{eq:mask}), the gradient of loss function $\mathcal{L}$ with respect to $\mathbf{\widetilde w}$ is written as
\begin{equation}
\label{eq:w_gradient}
\frac{\partial \mathcal{L}}{\partial \mathbf{\widetilde w}} = \frac{\partial \mathcal{L}}{\partial \mathbf{w}} \odot \mathbf{m}
\end{equation}

As $\mathbf{m}$ is a sparse tensor, many elements of ${\partial \mathcal{L}}/{\partial \mathbf{\widetilde w}}$ are zero, indicating that these masked variables are not updated. Even though network connections that contribute less to precision can be ignored in the current global structure, they may not be negligible in future global structures.
As a result, temporary zero masks do not mean unimportant, and it is worth keeping training for these zero-masked weight variables. In this work, the gradient of loss function $\mathcal{L}$ with respect to a weight variable $\mathbf{\widetilde w}$ is redefined as
\begin{equation}
\label{eq:weight_redefine}
\frac{\partial \mathcal{L}}{\partial \mathbf{\widetilde w}} \vcentcolon= \frac{\partial \mathcal{L}}{\partial \mathbf{w}}
\end{equation}
where ${\partial \mathcal{L}}/{\partial \mathbf{w}}$ is obtained through back-propagation. Similar to \cite{guo2016dynamic,he2018soft}, even if some weight variables are temporarily zero-masked during training, they will update later.

\subsection{Gradient redefinition for mask variables}
\label{sec:mask_ste}

Due to the derivatives of a unit step function are mostly zero, mask variables can not update using gradient-based optimization methods. To solve this problem, we investigate the fundamental principle of using STE for network pruning. Let us analyze a realizable case that uses the following loss function to update mask variables $\mathbf{\tilde{m}}$. 
\begin{equation}
\label{eq:ste_l2_loss}
\min_{\mathbf{\tilde{m}}} \ell(\mathbf{\tilde{m}}) = \frac{1}{2} (H(\mathbf{\tilde{m}})-\mathbf{m^{*}})^{2}
\end{equation}
where $\mathbf{H(\tilde{m})}$ and $\mathbf{m^{*}}$ represent the binarized mask values and optimal mask values, respectively. Note that the optimal solution of mask variables $\mathbf{\tilde{m}}$ is a region rather than a point. The gradient of $\mathbf{\tilde{m}}$ is expressed as 
\begin{equation}
\label{eq:ste_l2_gradient}
\frac{\partial \ell}{\partial \mathbf{\tilde{m}}} = \frac{\partial H(\mathbf{\tilde{m}})}{\partial \mathbf{\tilde{m}}} (H(\mathbf{\tilde{m}})-\mathbf{m^{*}})
\end{equation}

\begin{figure*}[!th]
\centering
\subfigure[ReLU STE]{\includegraphics[height=3.0cm]{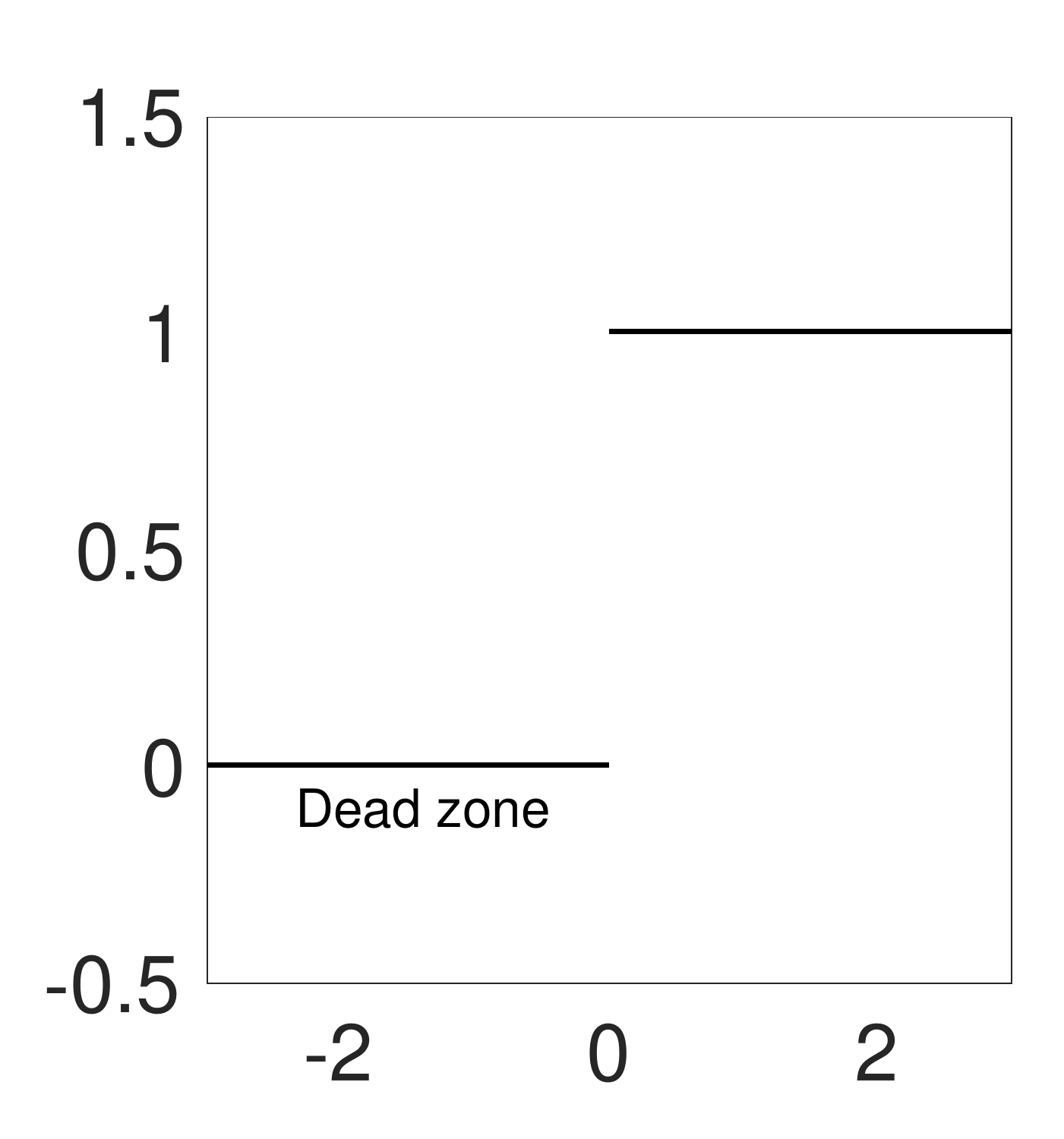}\label{fig:bi_relu}}
\subfigure[Clipped ReLU STE]{\includegraphics[height=3.0cm]{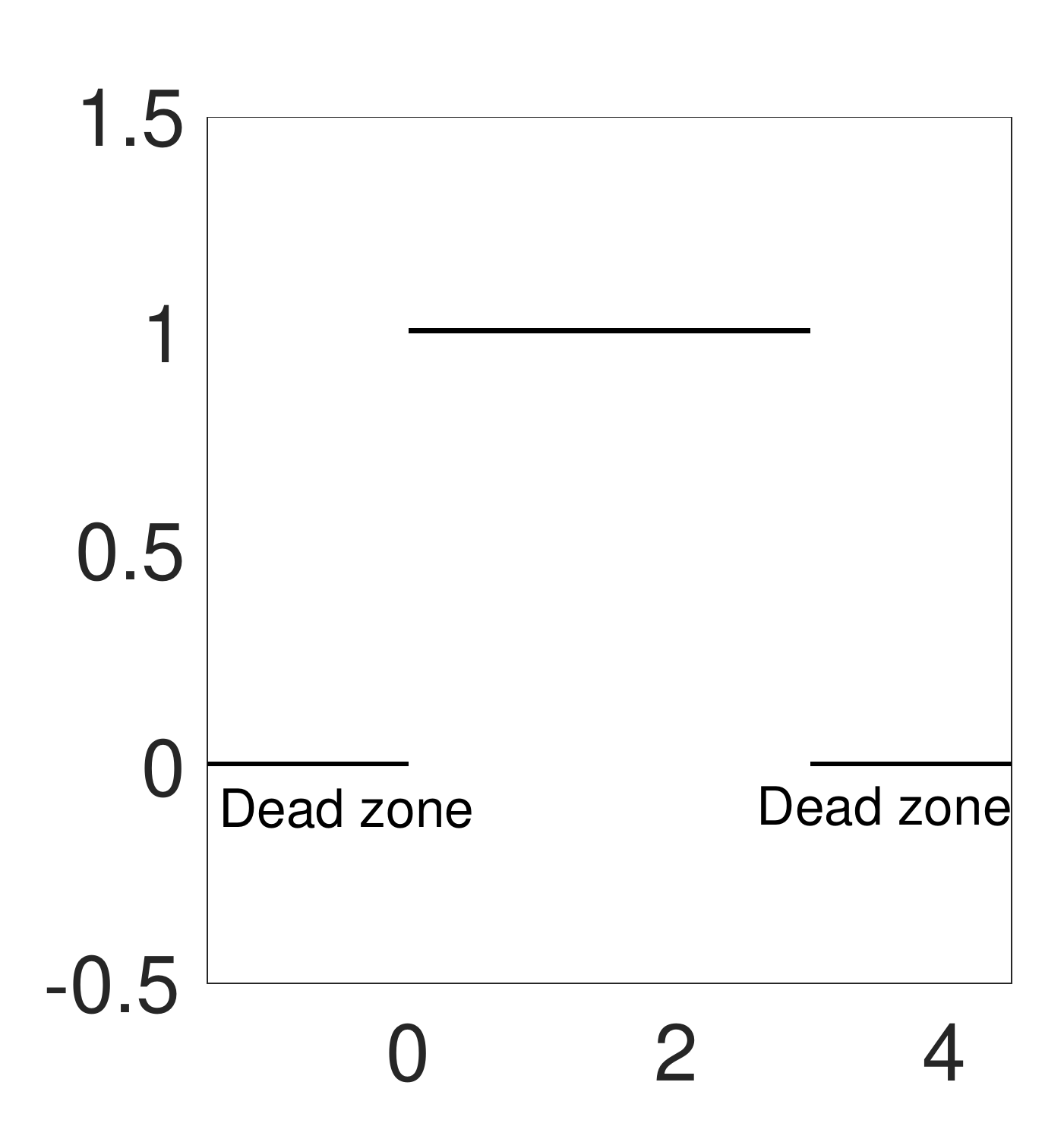}\label{fig:bi_crelu}}
\subfigure[Leaky ReLU STE]{\includegraphics[height=3.0cm]{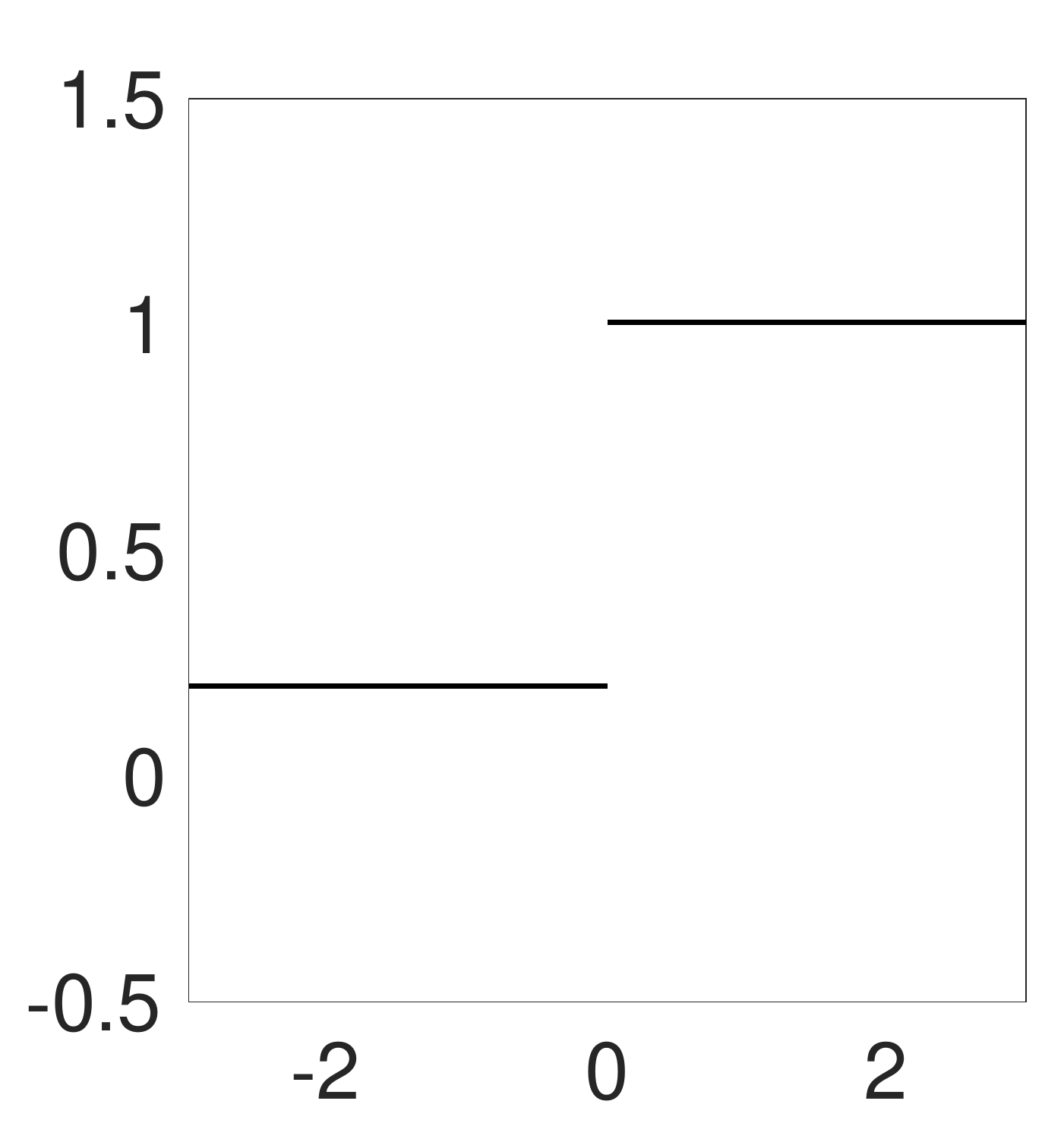}\label{fig:bi_lrelu}}
\subfigure[Softplus STE]{\includegraphics[height=3.0cm]{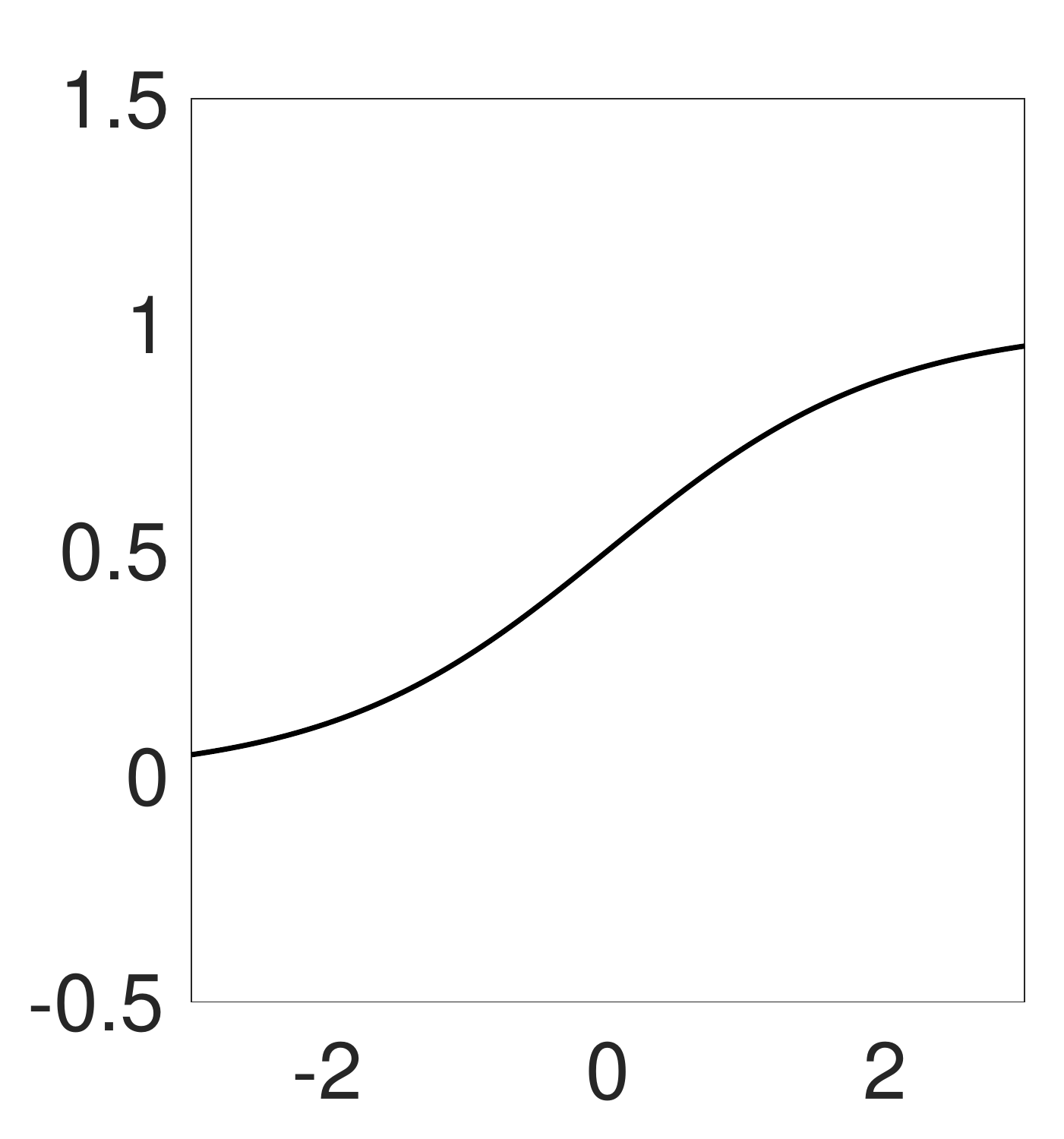}\label{fig:bi_softplus}}
\subfigure[Identity STE]{\includegraphics[height=3.0cm]{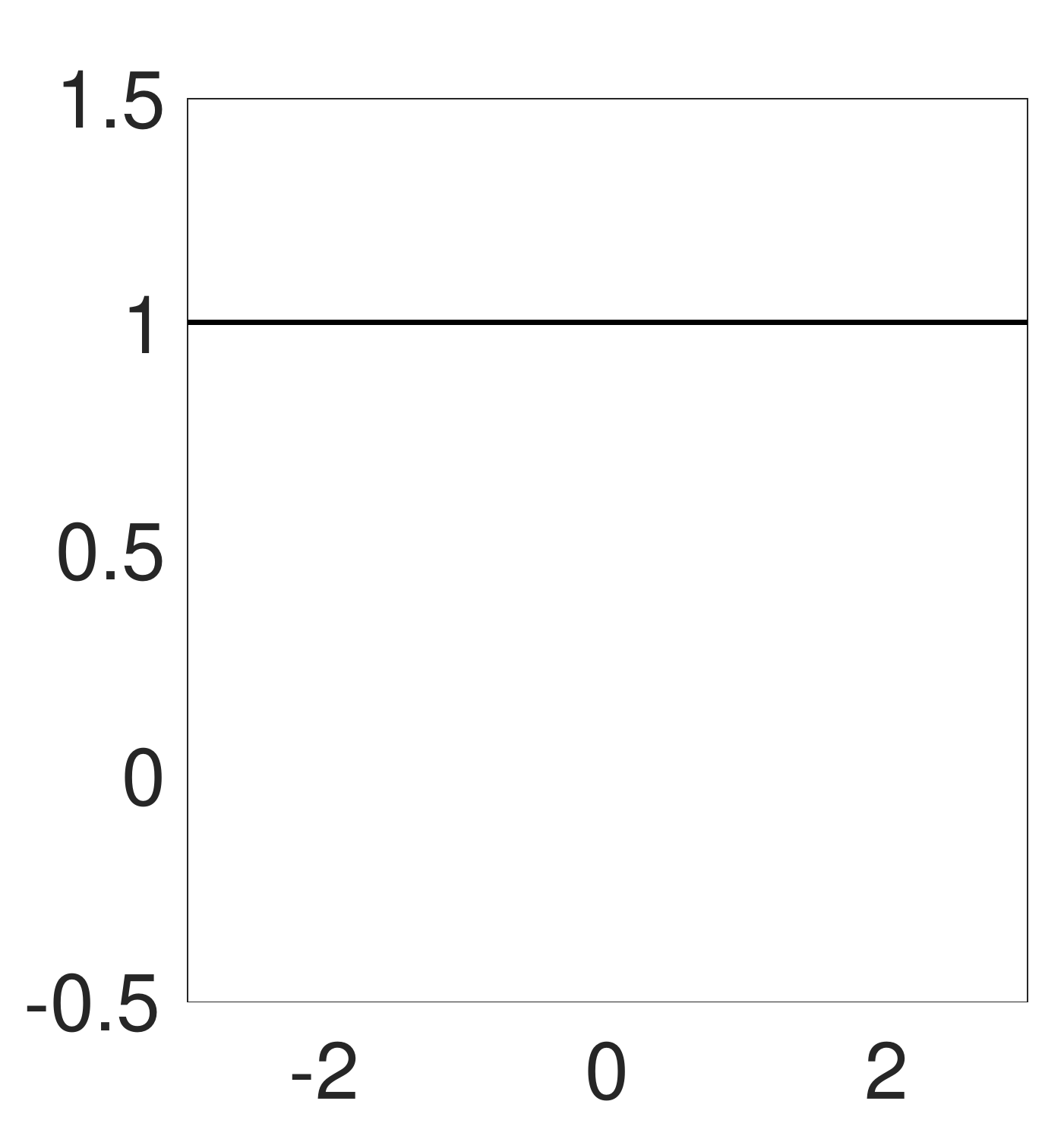}\label{fig:bi_identity}}
\caption{Proxy gradients of five existing STEs used for binarization relaxation. The X-axis and Y-axis represent mask variables $\mathbf{\tilde{m}}$ and proxy gradients ${\partial H(\mathbf{\tilde{m}})}/{\partial \mathbf{\widetilde m}}$, respectively. The dead zones for ReLU and Clipped ReLU STEs are marked.}
\label{fig:STEs}
\end{figure*}

As $H(\tilde{m}_{i})$ and $m^{*}_{i}$ are binary values, there are three possible values (\ie, 0, -1, and +1) for their difference $H(\tilde{m}_{i})- m^{*}_{i}$. Hence, the gradient of $\tilde{m}_{i}$ is expressed as
\begin{equation}
\label{eq:ste_l2_gradient_0}
\frac{\partial \ell}{\partial \tilde{m}}_{i} = 
\left\{
\begin{array}{l} 
\frac{\partial H(\tilde{m}_{i})}{\partial \tilde{m}_{i}} \cdot ~~0~~, \;\; if\;\; H(\tilde{m}_{i})= m^{*}_{i}\\
\frac{\partial H(\tilde{m}_{i})}{\partial \tilde{m}_{i}} \cdot (-1), \;\; if\;\; \tilde{m}_{i}\leq 0 ~and~ m^{*}_{i}=1 \\
\frac{\partial H(\tilde{m}_{i})}{\partial \tilde{m}_{i}} \cdot (+1), \;\; if\;\; \tilde{m}_{i}>0 ~and~ m^{*}_{i}=0
\end{array}
\right.
\end{equation}

When STE is used to relax the binarization, $\slfrac{\partial H(\tilde{m}_{i})}{\partial \tilde{m}_{i}}$ is replaced by proxy gradients. Correct proxy gradients should move $\tilde{m}_{i}$ towards their optimal values during network pruning. The gradient should be zero when the optimal value is reached, indicating no further update. Based on the mechanism of gradient descent optimizers (\ie, a variable is adjusted in the opposite direction of its gradient), $\tilde{m}_{i}$ should move towards the negative direction of proxy gradients. Consequently, when STE is used for network pruning, positive values of $\slfrac{\partial H(\tilde{m}_{i})}{\partial \tilde{m}_{i}}$ in Eq. (\ref {eq:ste_l2_gradient_0}) ensure mask variables converge at their minima. This fundamental principle is derived without any assumptions, and is just based on the mechanism of gradient descent optimizers. In contrast, a normal distribution is assumed for inputs in \cite{xiao2019autoprune}.

Figure \ref{fig:STEs} plots proxy gradients of the existing five STEs. We can see that three STEs (\ie, Leaky ReLU, Softplus, and identity) can satisfy the fundamental principle of positive proxy gradients. In this work, the Identity STE is selected for simplicity. Since its proxy gradient is always 1 as shown in Figure \ref{fig:bi_identity}, Eq. (\ref{eq:ste_l2_gradient_0}) is expressed as 

\begin{equation}
\label{eq:ste_l2_gradient_1}
\frac{\partial \ell}{\partial \tilde{m}}_{i} = 
\left\{
\begin{array}{l}
~0~, \;\;\;\;\; if\; H(\tilde{m}_{i})= m^{*}_{i}\\
-1, \;\;\;\;\;if\; \tilde{m}_{i}\leq 0 ~and~ m^{*}_{i}=1 \\
+1, \;\;\;\;\;if\; \tilde{m}_{i}>0 ~and~ m^{*}_{i}=0
\end{array}
\right.
\end{equation}

As indicated in Eq. (\ref{eq:ste_l2_gradient_1}), when $\tilde{m}_{i}$ reaches its optimal value, the gradient is zero. As a result, $\tilde{m}_{i}$ does not update further. When $\tilde{m}_{i}\leq 0$ and $m^{*}_{i}=1$, the gradient of -1 pushes $\tilde{m}_{i}$ to be more positive. When $\tilde{m}_{i}>0$ and $m^{*}_{i}=0$, the gradient of +1 pushes $\tilde{m}_{i}$ to be more negative. Thus, mask variables and network connectivity (from 1 to 0 or vice versa) can update during training. The Identity STE ensures that mask variables move towards and finally stabilize at their optimal values. According to Eq. (\ref {eq:step}) and Figure \ref{fig:bi_identity}, we obtain
\begin{equation}
\label{eq:step_redefine3}
\frac{\partial \mathbf{m}}{\partial \mathbf{\widetilde m}} = \frac{\partial H(\mathbf{\tilde{m}})}{\partial \mathbf{\widetilde m}}=1
\end{equation}
which indicates that the differential of $\tilde{m}$ is redefined as that of $m$. Thus, the gradient of loss function $\mathcal{L}$ with respect to mask variables $\mathbf{\widetilde m}$ is redefined as
\begin{equation}
\label{eq:step_redefine}
\frac{\partial \mathcal{L}}{\partial \mathbf{\widetilde m}} = \frac{\partial \mathcal{L}}{\partial \mathbf{m}} \frac{\partial \mathbf{m}}{\partial \mathbf{\widetilde m}} := \frac{\partial \mathcal{L}}{\partial \mathbf{m}} = \frac{\partial \mathcal{L}}{\partial \mathbf{w}} \odot \mathbf{\widetilde w}
\end{equation}


\subsection{Mask gradient normalization}
\label{sec:mask_norm}

\begin{figure*}[!ht]
\centering
\subfigure[Average in layers]{\includegraphics[height=3.4cm]{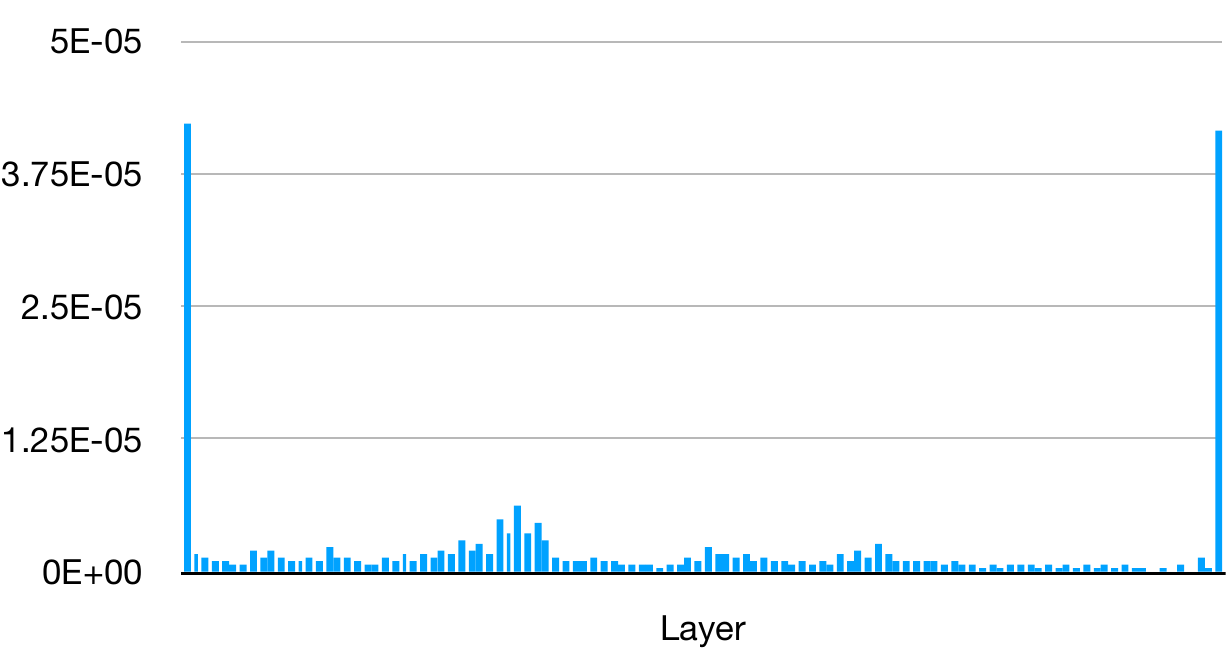}\label{fig:grad_layers}}
\hspace{0.5cm}%
\subfigure[Layer-1]{\includegraphics[height=3.4cm]{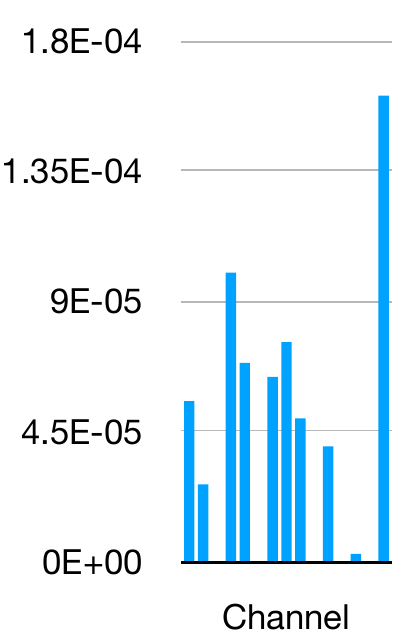}\label{fig:grad_layer0}}
\hspace{0.5cm}%
\subfigure[Layer-105]{\includegraphics[height=3.4cm]{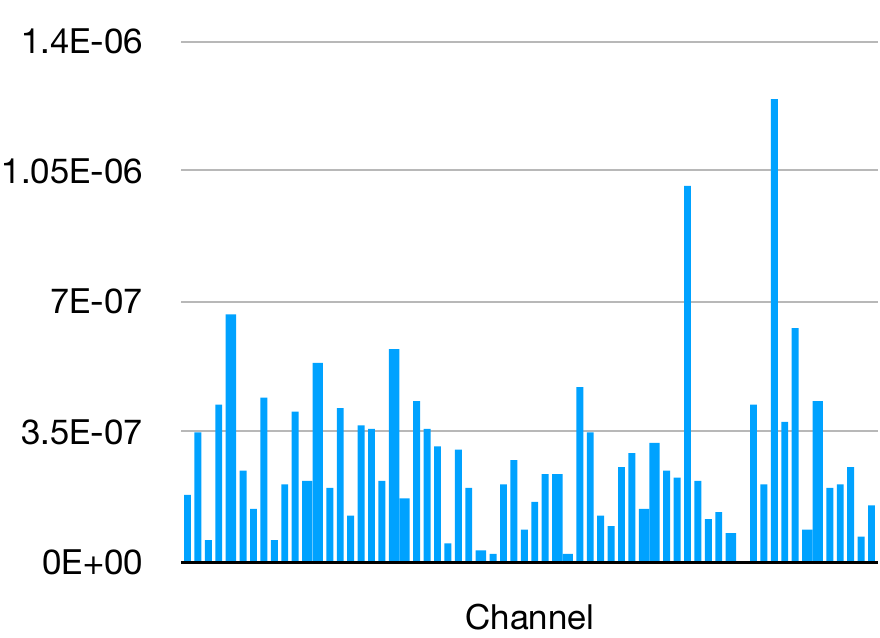}\label{fig:grad_layer46}}
\caption{Mask gradient magnitudes of pre-trained ResNet-110. (b) and (c) show mask gradient magnitudes of channels in layer-1 and layer-105, respectively.}
\label{fig:grad}
\end{figure*}

We observe a wide range of mask gradient magnitudes in various layers and channels of common neural networks. To visualize this observation, a pre-trained ResNet-110 is used as an example to plot mask gradient magnitudes in Figure \ref{fig:grad}. In Figure \ref{fig:grad_layers}, layer-wise results show that the average mask gradient magnitude of the first and last layers (\eg, $10^{-5}$) is much higher than other layers (\eg, $10^{-7}$). Then, channel-wise results of two typical layers in Figures \ref{fig:grad_layer0} and \ref{fig:grad_layer46} show that mask gradient magnitudes fluctuate greatly between channels. 

The influence of a wide range of mask gradient magnitudes is analyzed below. Let us review the training process of mask variables during network pruning. Through binarization (\ie, Eq. (\ref{eq:step})), a positive mask variable means a mask state of 1, indicating that there is a network connection. If this network connection is pruned, the binary mask state should become 0. According to Eq. (\ref{eq:step}), the final value of this mask variable should be zero or negative. Assume two mask variables are initialized to the same positive value and eventually become zero. That is, two initially established network connections are eliminated after pruning. Under the same learning rate for all layers, the mask variable with a smaller gradient magnitude requires more training iterations to update until convergence. Therefore, it is difficult for mask variables with a wide range of mask gradient magnitudes to converge to their optimal solution, thereby degrading network pruning performance. To mitigate this problem, we propose to normalize mask gradients on different features. Mask gradients of each feature are normalized to the unit variance in each mini-batch. As shown in Eq. (\ref{eq:grad_norm}), mask gradients obtained through back-propagation are divided by a gradient scale $s$.
\begin{equation}
\label{eq:grad_norm}
\begin{split}
& Norm(\frac{\partial \mathcal{L}}{\partial \mathbf{m}_{j}}) = \slfrac{\frac{\partial \mathcal{L}}{\partial \mathbf{w}_{j}} \odot \mathbf{\widetilde w}_{j}}{(s + \epsilon)}, \;\;where \\
& s = \sqrt{\slfrac{\sum_{\mathbf{x}_{b} \in \mathcal{B}} \sum_{w_{k} \in \mathbf{w}_{j}} (\frac{\partial \mathcal{L}(\mathbf{x}_{b})}{\partial w_{k}} \odot \widetilde w_{k})^{2}}{|\mathcal{B}|\cdot|\mathbf{w}_{j}|}}
\end{split}
\end{equation}
where $\mathcal{B}$ and $|\mathcal{B}|$ denote the sampled mini-batch data and batch size, respectively. $\mathbf{w}_{j}$ and $|\mathbf{w}_{j}|$ represent the weight and weight number of the $j$-th feature, respectively. $\epsilon$ is a small positive constant to avoid division by zero. Since it is not a zero-mean normalization, the sign of each mask gradient does not change, so the analysis derived in the previous subsection is still valid. Because the same gradients are produced on all masks by the connectivity decay introduced in the next subsection, the gradient normalization excludes mask gradients caused by the connectivity decay (\ie, normalization in Eq. (\ref{eq:grad_norm}) processes the gradients produced by the first and third terms of Eq. (\ref{eq:loss})). 

\subsection{Connectivity decay for sparse connectivity learning}

Through the proposed weight re-parameterization, the information of network connectivity is completely represented by the elements in the binary mask $\mathbf{m}$. The degree of network connectivity is equal to the sum of all element values in $\mathbf{m}$. We incorporate the degree of connectivity into an objective function to optimize network connectivity. The training objective function is expressed as
\begin{equation}
\label{eq:connectivity}
\mathcal{L} = \mathcal{C} + \lambda_{1} \cdot \sum_{l=0}^{L} \sum_{i=0}^{|\mathbf{m}^{(l)}|} m_{i}^{(l)}
\end{equation}
where $\mathcal{C}$ is the criteria to measure performance loss, $\mathbf{m}^{(l)}$ is the mask of $l$-th layer, $i$ is the element index, $|\mathbf{m}^{(l)}|$ is the total number of elements in $l$-th layer, and $L$ is the total number of layers. $\lambda_{1}$ is a hyper-parameter for connectivity decay. As the second term includes the information of network connectivity, so during training this term attenuation indicates connectivity decay. Connectivity decay can be viewed as a way of $L_{0}$ regularization of weight. There are two gradients that affect the training of mask variables. One gradient is due to the connectivity decay, which pushes mask variables moving towards the negative infinity direction all the time. The other gradient is due to the performance loss minimization, which usually pushes mask variables moving towards the positive infinity direction according to the significance of network connectivity. When training converges, both gradients on the mask variables will reach equilibrium at important network connections. Therefore, an optimized neural network with an expected level of sparsity is learned and determined through training.
\subsection{Proposed SCL algorithm}

Algorithm \ref{algorithm} describes the details of our proposed automatic sparse connectivity learning method. Through repeatedly calculating gradients of weight variables and mask variables, this algorithm updates them through stochastic gradient descent (SGD).
When the network pruning process is complete, a sparse neural network is obtained. The well-trained sparse weights are calculated as
\begin{equation}
\label{eq:sparseweights}
\mathbf{w}^{*} = \mathbf{\widetilde w} \odot H(\mathbf{\widetilde m})
\end{equation}

\begin{algorithm}[!htb]
  \caption{Automatic Sparse Connectivity Learning}
  \label{algorithm}
  
  \begin{flushleft}
  \textbf{Input: } A training dataset $\mathcal{D}$, a $L$-layer neural network, weight variables $\mathcal{W}=\{\widetilde{\textbf{w}}^{(l)}\}_{l=1}^{l=L}$, mask variables $\mathcal{M}=\{\widetilde{\textbf{m}}^{(l)}\}_{l=1}^{l=L}$, a connectivity decay coefficient $\lambda_{1}$, an $L_{2}$ regularization coefficient $\lambda_{2}$, and the total number of training epochs $T$. \\
  \textbf{Output: } Well-trained sparse weights  $\mathcal{W}^{*}=\{{\textbf{w}^{(l)}}^{*}\}_{l=1}^{l=L}$. \\
  \textbf{Initialization: } Initial values of mask variables are positive. Initial values of weight variables are random according to \cite{he2015delving}.
  \end{flushleft}
  
  \begin{algorithmic}[1]
  \REPEAT
  \STATE Sample a mini-batch from $\mathcal{D}$ as input data.
  \STATE \textbf{Forward pass: }\\
  First, calculate all the masks using Eq. (\ref{eq:step}).
  Next, compute all weights using Eq. (\ref{eq:mask}). Then, Calculate all layers like a normal neural network with weights and sampled input.
  
  \STATE \textbf{Backward pass: }\\
  First, calculate weight gradients using Eq. (\ref{eq:weight_redefine}). Next, relax mask gradients through the Identity STE using Eq. (\ref{eq:step_redefine}). Then, normalize mask gradients using Eq. (\ref{eq:grad_norm}).
  
  \STATE \textbf{Update weights: }\\
  Use gradients obtained from the backward pass to update all weight variables and mask variables via SGD.
  
  \UNTIL{$T$ epochs complete}

  \STATE Compute the well-trained sparse weights using Eq. (\ref{eq:sparseweights}) and output a sparse neural network.
  \end{algorithmic}
\end{algorithm}

\subsection{Comparison of SCL with existing automatic pruning}

It is necessary to comprehensively compare our SCL algorithm with existing automatic pruning methods in the literature (\ie, \cite{louizos2017learning,kang2020operation,herrmann2020channel,huang2018data,xiao2019autoprune}). Therefore, conceptual comparison and experimental results discussion of \cite{louizos2017learning,huang2018data,kang2020operation,xiao2019autoprune} will be provided in Sections \ref{sec:sparse_compareCIFAR} and \ref{sec:sparse_compareIMAGE}. 
Since \cite{herrmann2020channel} does not report proper experimental results to compare, we only discuss their algorithm differences as below. Compared with \cite{herrmann2020channel}, SCL has the following differences. First, the way of dealing with the non-differentiation problem of discrete masks is different. SCL uses a deterministic STE to estimate the gradient of discrete masks, whereas \cite{herrmann2020channel} uses the Gumbel-Softmax trick to relax the discrete masks to a continuous form. Second, the way to reduce the computational workload is different. \cite{herrmann2020channel} focuses on conditional computation using dynamic inference, so masks vary with network inputs. Besides, since the efficiency of data handling is greatly affected, the conditional computational graph in \cite{herrmann2020channel} is not friendly to hardware accelerators. In contrast, SCL focuses on network pruning that is static during inference, so the computational graph is fixed after training. Therefore, the feature of static computation reduction in SCL is more appropriate than \cite{herrmann2020channel} for hardware acceleration.








\section{Baseline network architectures and Experimental Setups}
\label{sec:baseline}

We conduct the experiments on four NVIDIA TITAN XP GPUs using PyTorch\footnote{\url{https://pytorch.org}}.
We choose VGGs \cite{simonyan2014very}, ResNets \cite{he2016deep}, DenseNets \cite{huang2017densely}, and EfficientNets \cite{tan2019efficientnet} as our baseline CNN architectures, because most of the state-of-the-art neural networks are based on them. ResNets and DenseNets are outstanding due to their high accuracy and fewer number of trainable parameters. EfficientNet is a lightweight high-accuracy convolutional neural network architecture with much fewer parameters. In CNN experiments, the proposed SCL technique is evaluated using the datasets of MNIST \cite{yannMNIST}, CIFAR \cite{krizhevsky2009learning}, and ImageNet (\textit{i.e.}, full data of ImageNet2012 classification) \cite{ILSVRC15}. In RNN experiments, SCL is evaluated on a language model using the WikiText-2 dataset \cite{merity2017pointer}.

For the experiments on the MNIST dataset, the baseline is a DenseNet-based network, where only fully connected layers are implemented (\ie, no convolutional networks involved) to evaluate the proposed SCL method. Then, two-dimension $28\times28$ image samples are flattened before being fed into the neural network. The input images are normalized using the channel mean and standard deviation. Within this baseline network, sixteen fully connected layers with a growth rate of 8 are applied to feature extraction, followed by a softmax function for object classification. For the experiments on the CIFAR dataset, convolutional networks are evaluated. The VGGs and ResNets are adapted from the codes\footnote{\url{https://github.com/Eric-mingjie/rethinking-network-pruning}} of \cite{liu2019rethinking}. The codes of DenseNets are slightly different from the description in the paper of \cite{huang2017densely}, which is referred to official codes\footnote{\url{https://github.com/liuzhuang13/DenseNet}}. For the experiments on the ImageNet dataset, we only evaluate SCL on VGG-16, ResNet-50\footnote{\url{https://github.com/pytorch/vision/tree/master/torchvision/models}}, and EfficientNet-B0\footnote{\url{https://github.com/lukemelas/EfficientNet-PyTorch}} due to the limitation of computational resources. The language model used in the RNN experiments consists of an encoder embedding, two LSTM layers, and a decoder embedding. As encoder and decoder embeddings are tied to improve the perplexity results \cite{inan2017tying}, we only account for one of the encoder/decoder embeddings in sparsity statistics. The vocabulary size and LSTM hidden layer size are 33,278, and 1,500, respectively. The language model used in the RNN experiments is adapted from a PyTorch word language model\footnote{\url{https://github.com/pytorch/examples/tree/master/word_language_model}}.

All the networks are trained by the SGD optimizer. We adopt the weight initialization method in \cite{he2015delving}, and batch normalization in \cite{ioffe2015batch} for fast training. The objective function of an \textit{L}-layer network is expressed as
\begin{equation}
\label{eq:loss}
\mathcal{L} = \mathcal{C} + \lambda_{1} \cdot \sum_{l=0}^{L} \sum_{i=0}^{|\mathbf{m}^{(l)}|} m_{i}^{(l)} + \lambda_{2} \cdot \sum_{l=0}^{L} \sum_{i=0}^{|\mathbf{\widetilde w}^{(l)}|} (\widetilde w_{i}^{(l)})^{2}
\end{equation}
where $\mathcal{C}$ is the cross-entropy loss for classification, $\lambda_{1}$ and $\lambda_{2}$ are the coefficients of connectivity decay and $L_{2}$ regularization, respectively. $|\mathbf{m}^{(l)}|$ and $|\mathbf{\widetilde w}^{(l)}|$ refer to the number of elements in $\mathbf{m}^{(l)}$ and $\mathbf{\widetilde w}^{(l)}$, respectively. Note that $\lambda_{1}$ and $\lambda_{2}$ are not designer-defined pruning criteria or layer-wise pruning hyper-parameters (\eg, pruning threshold or ratio), which are indispensable in existing pruning methods in the literature. In our proposed SCL method, the layer-wise sparsity is automatically determined by the learning algorithm itself, without the intervention of designers. In our SCL method, tuning $\lambda_{1}$ can adjust the trade-off between network sparsity and accuracy. Therefore, designers can adjust $\lambda_{1}$ to reach the desired sparsity-accuracy balance. For a given value of $\lambda_1$, we do not run full training iterations to check the resulting sparsity. Instead, we run a small portion (\eg, 10\%) of the full training iterations to see if the target sparsity can be achieved. In this way, a proper $\lambda_1$ for the target sparsity can be found with several attempts. To further reduce the trial time, we start with a larger value of $\lambda_1$, which leads to faster network pruning and hence shorter running time. Then, we decrease the value of $\lambda_1$ until the target sparsity is obtained. Therefore, the running time for tuning $\lambda_1$ is not huge. A similar process for tuning $\lambda_{1}$ is applied to other datasets and baseline architectures in the experiments of Section \ref{sec:experiments}. To make a concise comparison with existing works, we list the experimental results in terms of sparsity, the number of parameters, FLOPs, and accuracy in Tables \ref{tab:resnet20}-\ref{tab:rnn}. Note $\lambda_{2}$ is the coefficient of $L_{2}$ regularization, which has nothing with network pruning. $L_{2}$ regularization is required to ensure an effective learning rate when batch normalization is applied \cite{van2017l2}.

For experiments on the MNIST and CIFAR datasets, mini-batch size and initial learning rate are set to 64, and 0.1, respectively. For experiments on the MNIST dataset, a simple training schedule is used, and mask variables do not update in the first and last 15 epochs. For experiments on the CIFAR datasets, mask variables do not update in the first 150 epochs to facilitate weight convergence. Then, mask variables are updated to obtain the corresponding sparse network connectivity. The learning rate is adjusted to 0.01 when the target sparsity is almost achieved. Finally, the mask variables continue to update for 20 epochs and achieve stable network connectivity. In order to achieve ultimate convergence stability, the weights continue to update for 80 epochs with a learning rate of 0.01 and then another 80 epochs with a learning rate of 0.001. Initial values of mask variables are positive. Weight variables are randomly initialized according to \cite{he2015delving} and trained from scratch. A larger $\lambda_{1}$ usually leads to less number of training epochs for realizing the same sparsity expectation. For experiments on the ImageNet dataset, all the models are pre-trained before sparse training for saving time. Mini-batch size and initial learning rate are set to 128 and 0.005, respectively. We use 90 epochs for sparse connectivity training and another 60 epochs to achieve stable training convergence. In the RNN experiments, we follow the training procedure of Distiller\footnote{\url{https://github.com/NervanaSystems/distiller/tree/master/examples/word_language_model}}.

\section{Experimental Results and Comparison}
\label{sec:experiments}

\subsection{Effect of STE and mask gradient normalization}
\label{sec:experiments_mapping}

An map fitting experiment has been designed to evaluate the effect of STE and mask gradient normalization. A three-layer fully connected network is built as a baseline input-output mapping. Each network layer consists of 64 neurons, followed by batch normalization and ReLU activation. In order to ensure output convergence, weights are randomly initialized according to \cite{he2015delving}. A set of inputs and mask variables are randomly selected and applied to this baseline. These inputs and mask variables follow a normal distribution with 0-mean and unit standard deviation. Thus, without any training, output results of the baseline are obtained as a benchmark. 

Next, the baseline architecture is reused to perform output fitting experiments, where inputs and weights are the same as the baseline, but mask variables are randomly initialized and then trained through gradient descent to match the benchmark (\ie, baseline outputs). These fitting experiments involve different STEs (\ie, proxy mask gradients) for binary mask relaxation with or without mask gradient normalization. In these experiments, the ideal fitting result is that after training, mask variables finally converge and produce the same outputs as the benchmark, indicating an accurate reproduction of the baseline input-output mapping. 

\begin{figure}[!t]
\centering
\includegraphics[height=5.5cm]{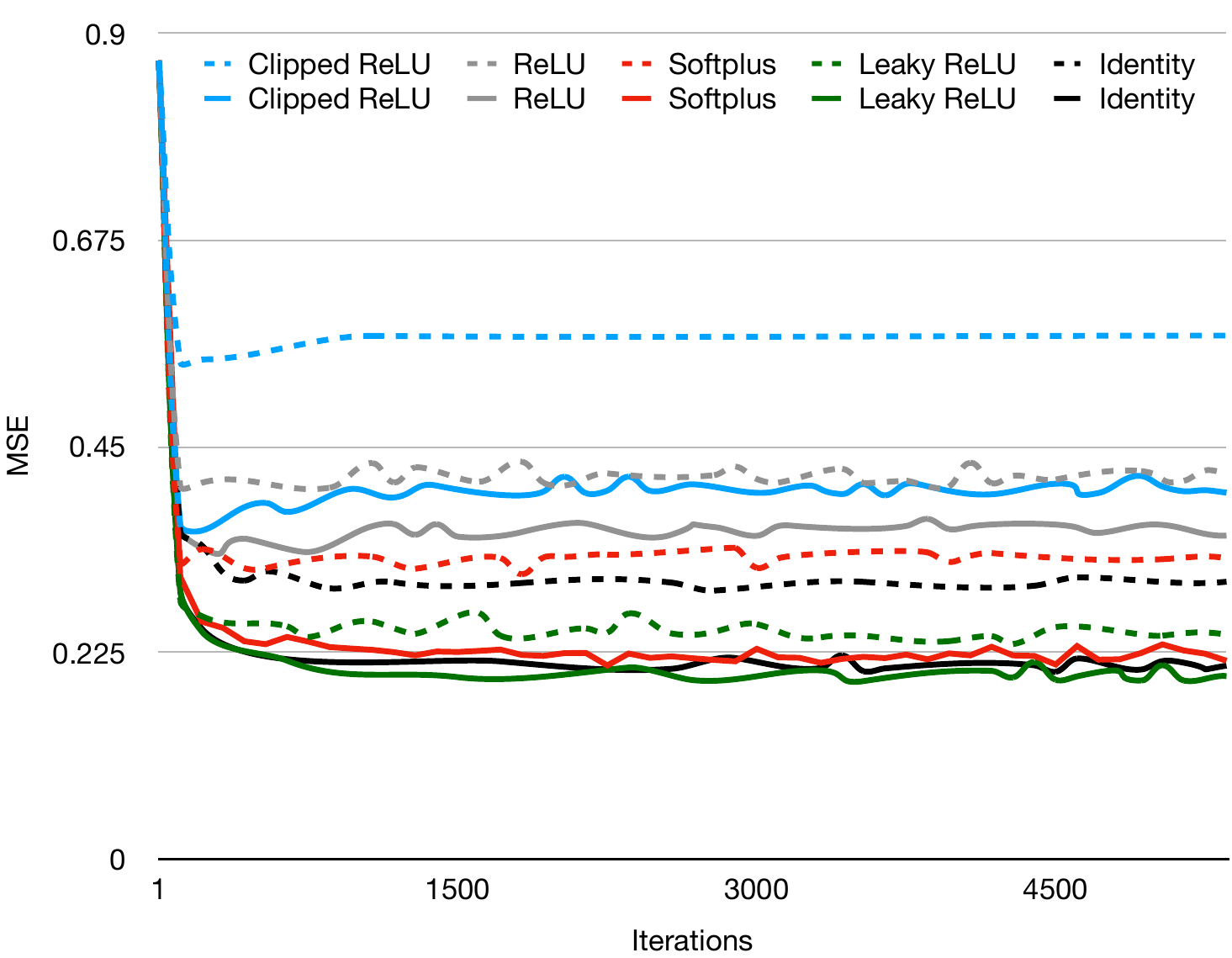}
\caption{Output fitting results with STEs for binary mask relaxation. Solid and dashed curves are with or without mask gradient normalization, respectively. }
\label{fig:mask_ste}
\end{figure}

Figure \ref{fig:mask_ste} shows the mean squared errors (MSEs) obtained from the above experiments. MSE tells how close the outputs of trained networks are to the benchmark. The Clipped ReLU STE leads to the worst fitting results, because its gradient is 0 when a mask variable is negative or larger than a threshold as shown in Figure \ref{fig:bi_crelu}. The results of ReLU STE are better, because it has fewer dead zones as shown in Figure \ref{fig:bi_relu}. Note that once a mask variable falls into a dead zone, there is no chance to get out. Compared with the Clipped ReLU and ReLU STEs, Leaky ReLU, Softplus, and Identity STEs achieve better and similar results. This observation is consistent with Eq. (\ref{eq:ste_l2_gradient_0}) and theoretical analysis in Section \ref{sec:mask_ste}. Positive proxy gradients of these three STEs guarantee a loss descent direction to push mask variables towards the minima. Mask gradient normalization reduces MSEs for all STEs, validating the necessity of normalizing mask gradients. 

\subsection{Sparse connectivity of fully connected network on MNIST}

To quickly evaluate the effectiveness of the proposed SCL method, we carried out experiments using a DenseNet-based network with the MNIST database, as described in Section \ref{sec:baseline}. As shown in Table \ref{tab:mnist}, a larger connectivity decay $\lambda_{1}$ corresponds to more sparse network connectivity. Even if $\lambda_{1}$ is set to 0, our experimental results show that the two gradients due to performance loss minimization and $L_{2}$ regularization (the first and third term in Eq. (\ref{eq:loss})) can push some mask variables to negative values, when their corresponding network connections do no contribute to accuracy. As a result, our SCL-induced neural network has a good sparsity of 34\%, and its object classification accuracy of 98.47\% outperforms the DenseNet-based baseline network (\ie, 98.35\%). From a system accuracy perspective, this means that the best network connection for this example is inherently sparse. In addition, when $\lambda_{1}$ is 0.03, our SCL-induced sparse networks can achieve a very high accuracy of above 98\% and sparsity of about 96.2\%. Note that in this fully connected layer experiment, the sparsity is equal to the reduction in FLOPs. As a result, the SCL-induced DenseNet-based network achieves a 96.2\% reduction in FLOPs (\ie, approximately $26.3 \times$ lower than the baseline with an accuracy loss of only 0.34\%). This experiment validates that our proposed SCL method supports effective network pruning on fully connected layer architectures.
\begin{table}[!t]
  \caption{Trained DenseNet-based networks on MNIST.}
  \label{tab:mnist}
  \centering
  \begin{tabular}{@{}l
                  S[table-format=6]
                  S[table-format=2.1]
                  S[table-format=2.2]@{}}
    \toprule
 {Scheme}  & {\# Param.} & {Sparsity}   &  {Accuracy}  \\
    \midrule
 {Baseline}            &    117152          &    0\si{\percent}  &  98.35\si{\percent}  \\
     \midrule
{$\lambda_{1}$ = 0}&        77422          &    34.0\si{\percent}  &   98.47\si{\percent} \\
{$\lambda_{1}$ = 0.01}&     10153          &   91.3\si{\percent}   &  98.24\si{\percent}  \\
{$\lambda_{1}$ = 0.03}&      4488          &   96.2\si{\percent}   &  98.01\si{\percent}  \\
{$\lambda_{1}$ = 0.08}&      1375          &   98.8\si{\percent}   &  94.64\si{\percent}  \\
{$\lambda_{1}$ = 0.1} &      252           &   99.8\si{\percent}   &  78.46\si{\percent}  \\
    \bottomrule
  \end{tabular}
\end{table}

\begin{figure}[!t]
\centering
\subfigure[$\lambda_{1}$ = 0.1]{\includegraphics[height=2cm]{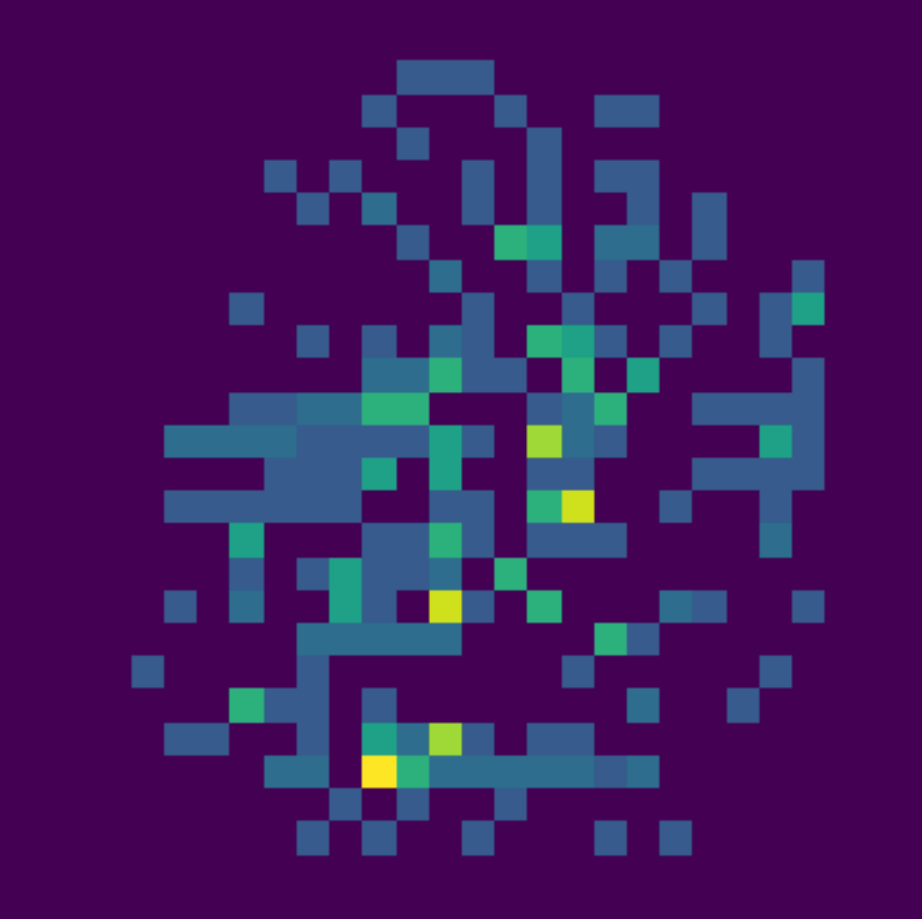}\label{fig:connection_visual.1}}
\hspace{0.2cm}%
\subfigure[$\lambda_{1}$ = 0.03]{\includegraphics[height=2cm]{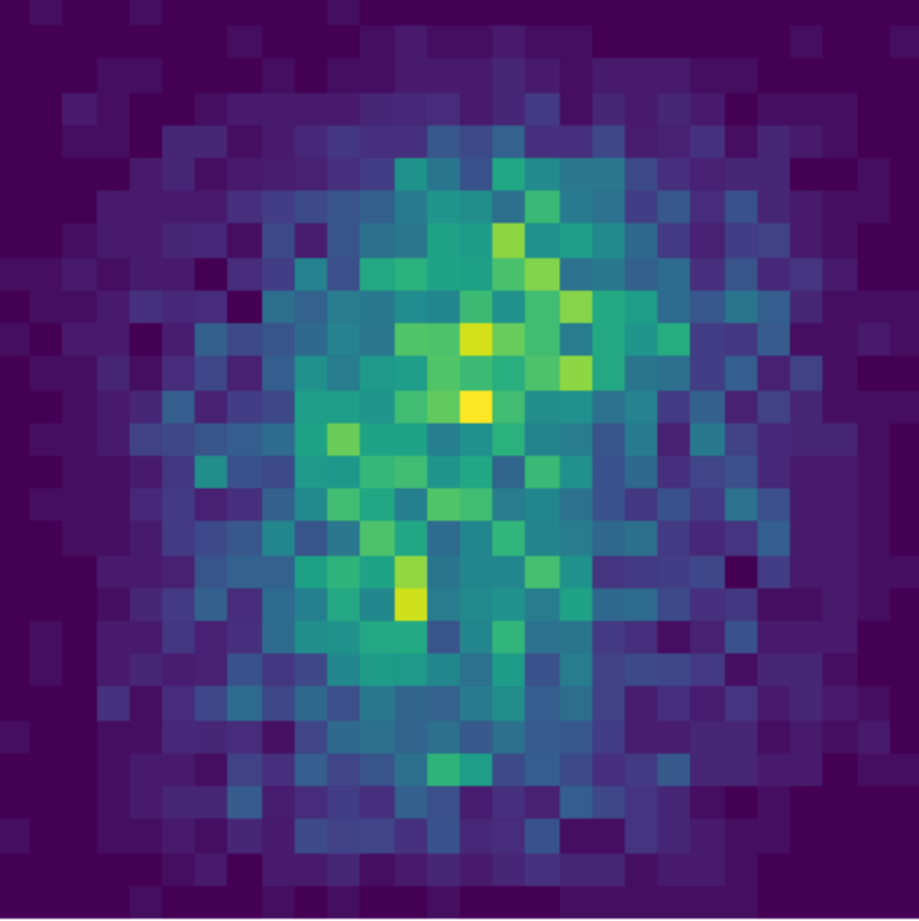}\label{fig:connection_visual.03}}
\hspace{0.2cm}%
\subfigure[$\lambda_{1}$ = 0.01]{\includegraphics[height=2cm]{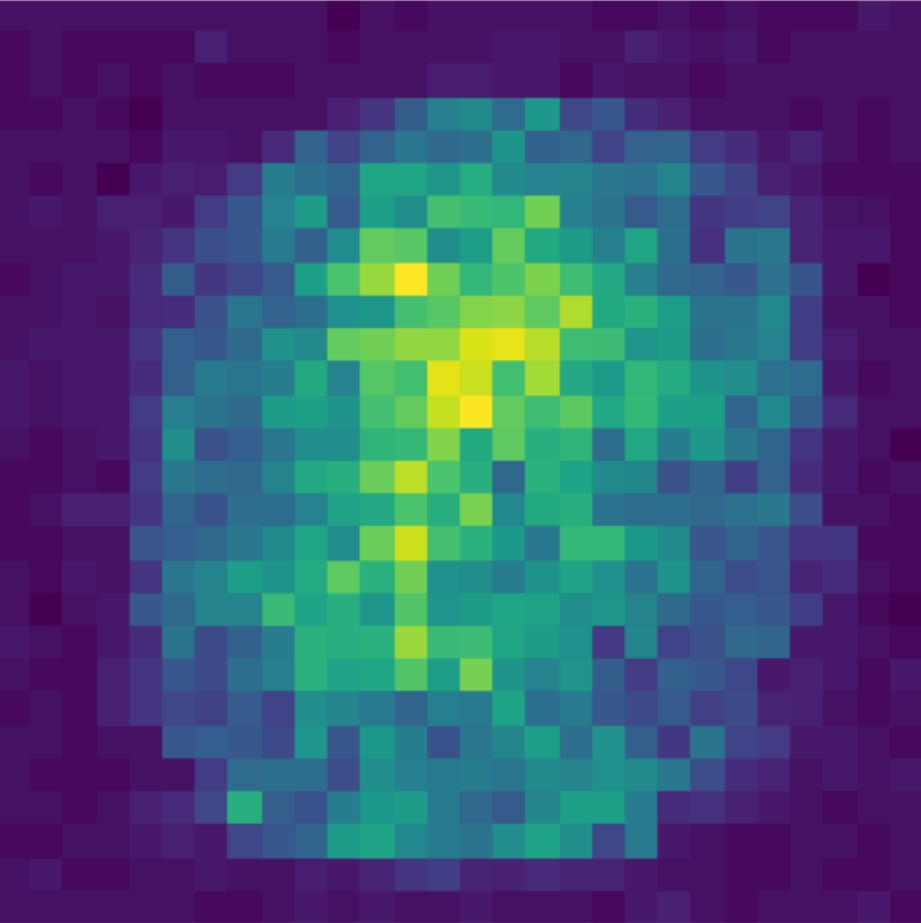}\label{fig:connection_visual.01}}
\caption{Visualization of learned network connectivity on MNIST.}
\label{fig:connection_visual}
\end{figure}

Based on these experimental results, we add together the binary connections that are directly connected to the input data and then normalize them. The normalized connections for three typical $\lambda_{1}$ values are illustrated in Figure \ref{fig:connection_visual}. It is clear that network connections are mainly concentrated in central areas, so our SCL training ignores edges. As a result, the pixels of digits are trained to be connected in the central area. When $\lambda_{1}$ is set to 0.1 (\textit{i.e.}, very high sparsity), only a few central pixels are connected as illustrated in Figure \ref{fig:connection_visual.1}. 

\subsection{Connectivity learning analysis on CIFAR-10}
\label{sec:sparse_analysis}

\begin{figure}[!t]
\centering
\subfigure[DenseNet-40 (overall $\thickapprox$ 10\%)]{\includegraphics[width=4.35cm]{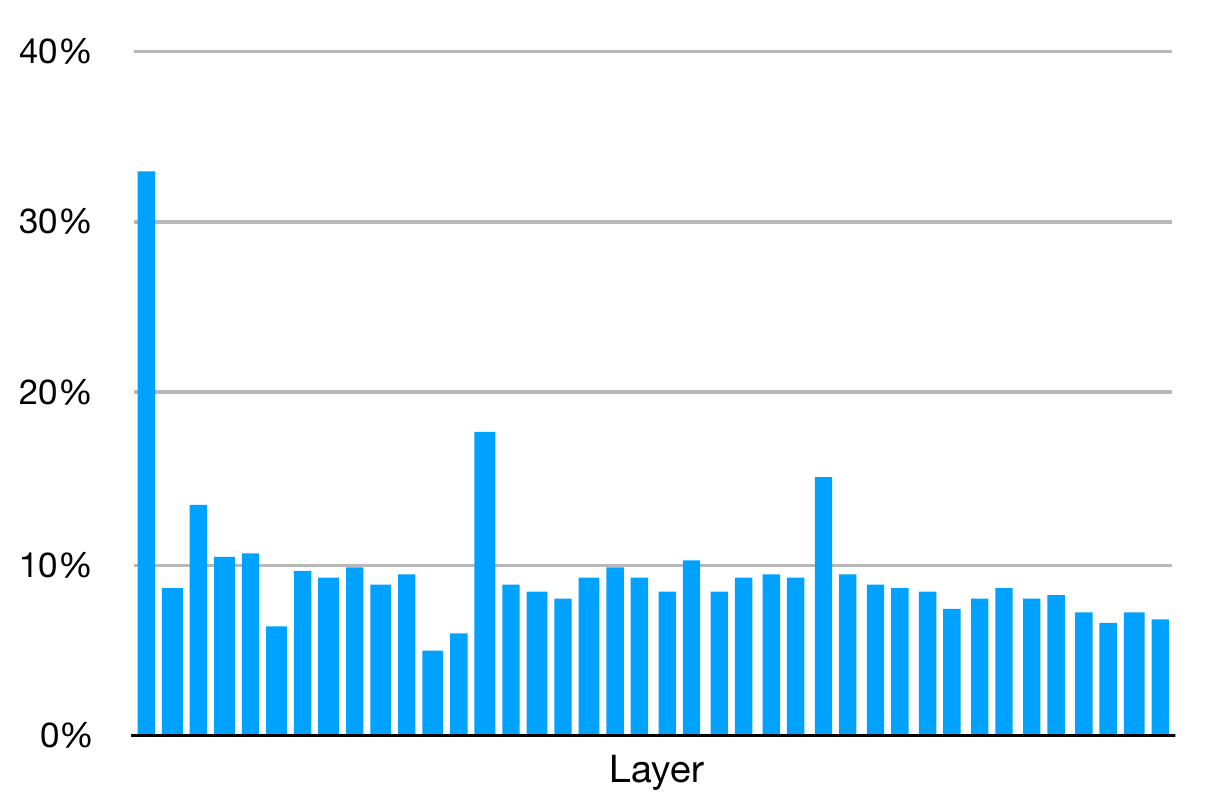}\label{fig:sparsity.densenet40.10}}
\subfigure[DenseNet-40 (overall $\thickapprox$ 30\%)]{\includegraphics[width=4.35cm]{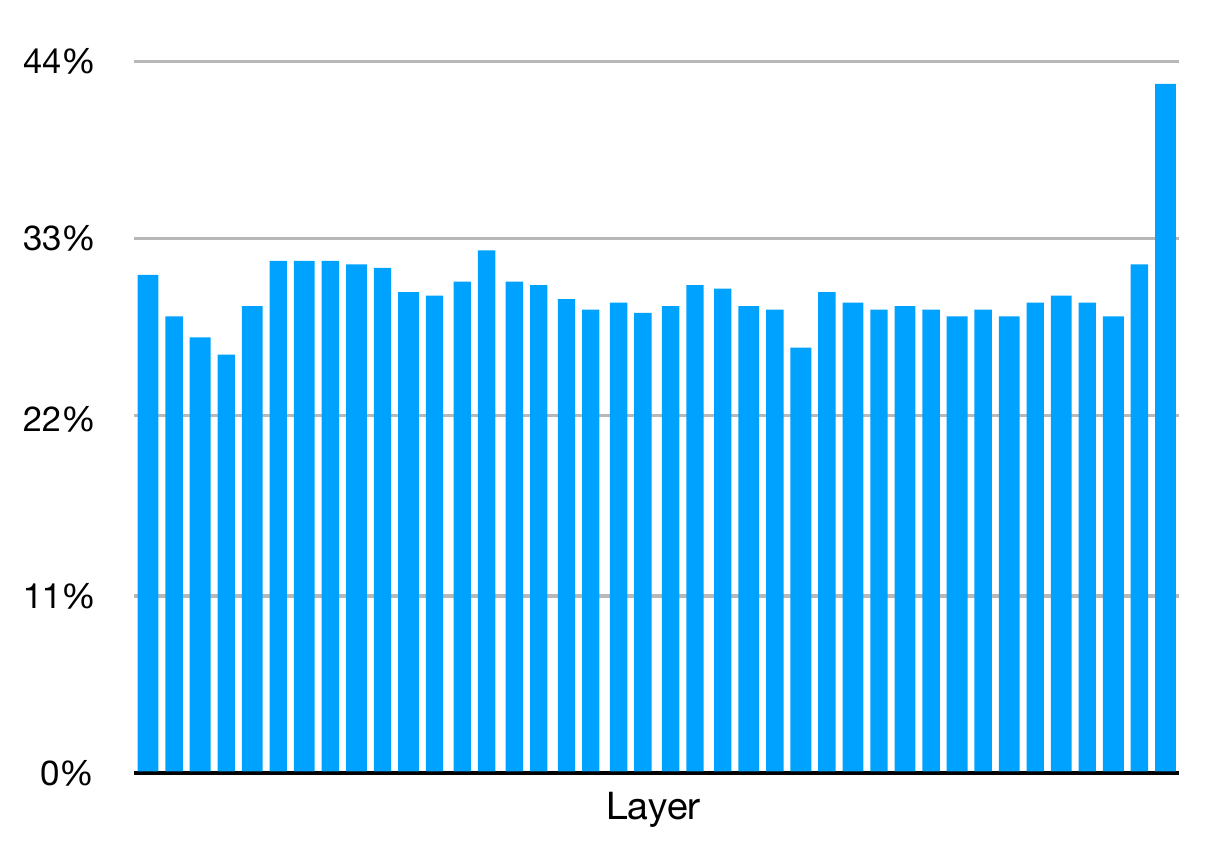}\label{fig:sparsity.densenet40.30}}
\subfigure[ResNet-110 (overall $\thickapprox$ 10\%)]{\includegraphics[width=4.35cm]{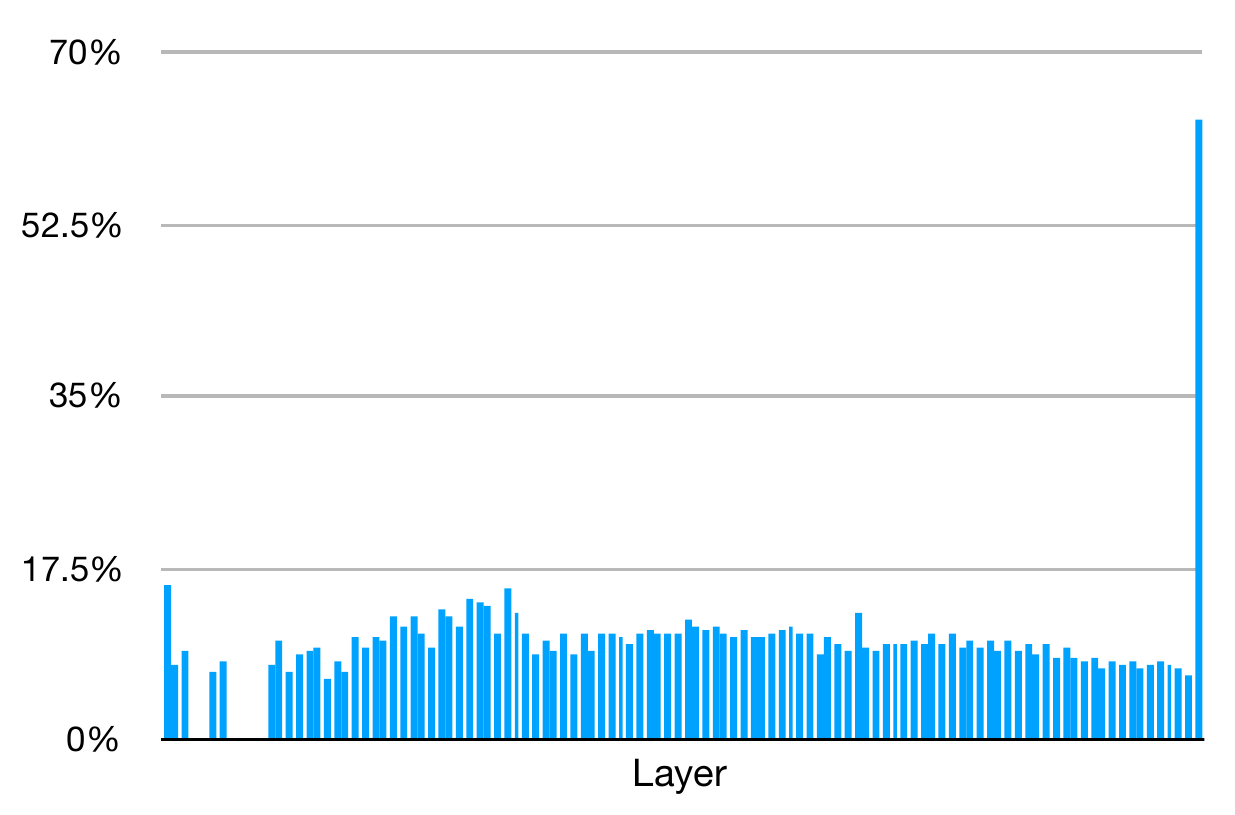}\label{fig:sparsity.resnet110.10}}
\subfigure[ResNet-110 (overall $\thickapprox$ 30\%)]{\includegraphics[width=4.35cm]{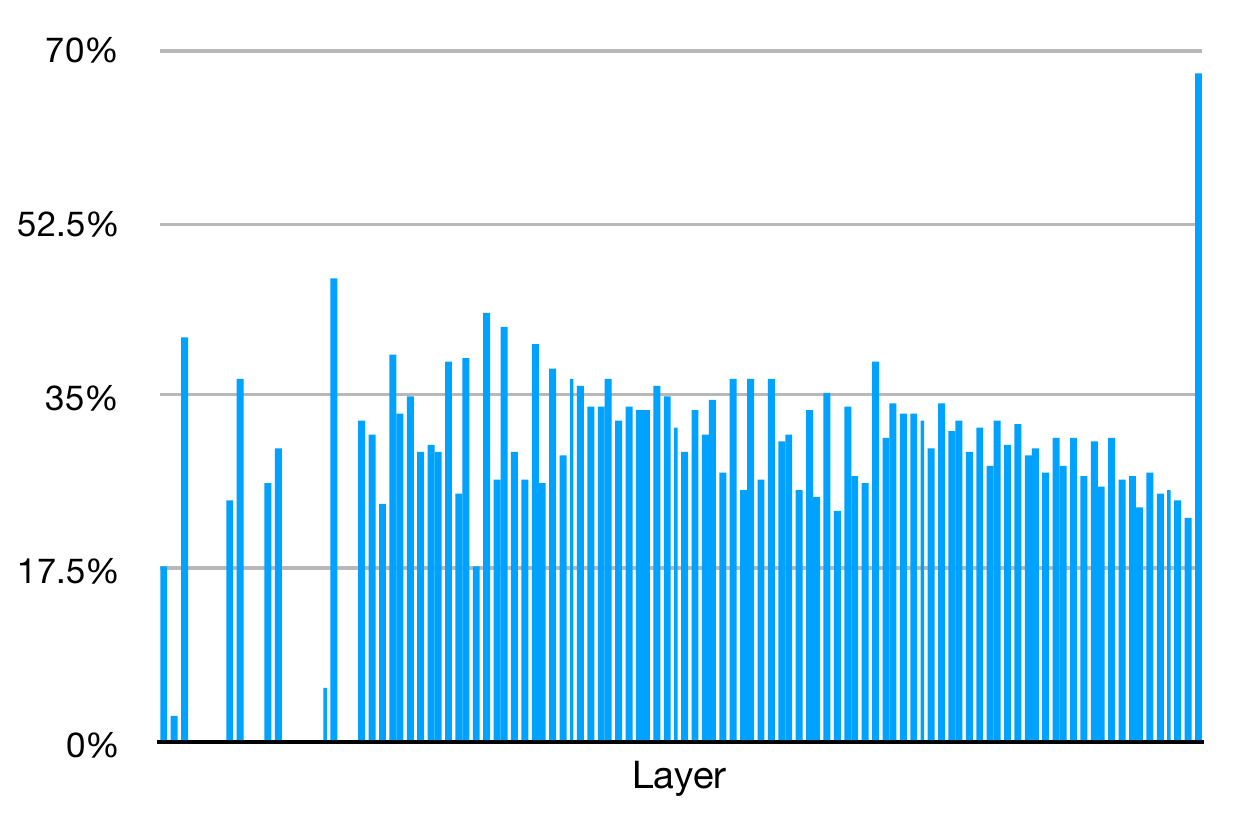}\label{fig:sparsity.resnet110.30}}
\caption{Connection density profiles for all layers in DenseNet-40 and ResNet-110 for different overall densities.}
\label{fig:sparsity}
\end{figure}

In order to demonstrate the automatically learned network connectivity, we report and discuss the density profiles (\ie, the percentage of remaining network connections) of several SCL-induced networks on the CIFAR-10 dataset. Figure \ref{fig:sparsity} shows that the proposed SCL method can automatically learn and determine the corresponding density profile for each network layer. In Figure \ref{fig:sparsity.densenet40.10}, there are three layers with relatively high densities, which are the layers located before the three dense blocks. Since it is most often reused in this low-density DenseNet network, the three important network layers should retain more connections. In Figure \ref{fig:sparsity.densenet40.30}, when this DenseNet network has a high targeted density of about 30\%, most of the network layers obtained by our SCL method have a similar density. Note that the density of the last network layer is completely different in Figure \ref{fig:sparsity.densenet40.10} and Figure \ref{fig:sparsity.resnet110.10}. This is due to different types of redundancy between DenseNet and ResNet structures. In DenseNet-40, the fact that previously extracted features are channel-wise concatenated to subsequent layers leads to 456 channel connections to the last layer. In ResNet-110, previously extracted features are element-wise added to subsequent layers, so the last layer contains only 64 channel connections and there is less redundancy. Therefore, DenseNet-40 removes most weights in the last layer, whose density is about 7\% in Figure \ref{fig:sparsity.densenet40.10}, while ResNet-110 retains most weights in the last layer, whose density is about 65\% in Figure \ref{fig:sparsity.resnet110.10}. In Figure \ref{fig:sparsity.resnet110.10} and \ref{fig:sparsity.resnet110.30}, since ResNet-110 is too deep, network connections in some layers are completely removed. In Figure \ref{fig:sparsity.resnet110.30}, even though 70\% of weights and some computational expensive shallow layers are removed, our SCL-induced network is superior to the baseline networks (94.82\% vs. 93.57\% in Table \ref{tab:resnet110}). In summary, we find that unlike conventional pruning methods which require designer-defined pruning criteria or hyper-parameter for each layer, our proposed SCL method can automatically learn and select important network connections for given baseline structures.

\begin{figure*}[!t]
\centering
\subfigure[VGG-16, comparisons with Frankle \etal \cite{frankle2018lottery}, and Li \etal \cite{li2016pruning}.]{\includegraphics[width=5cm]{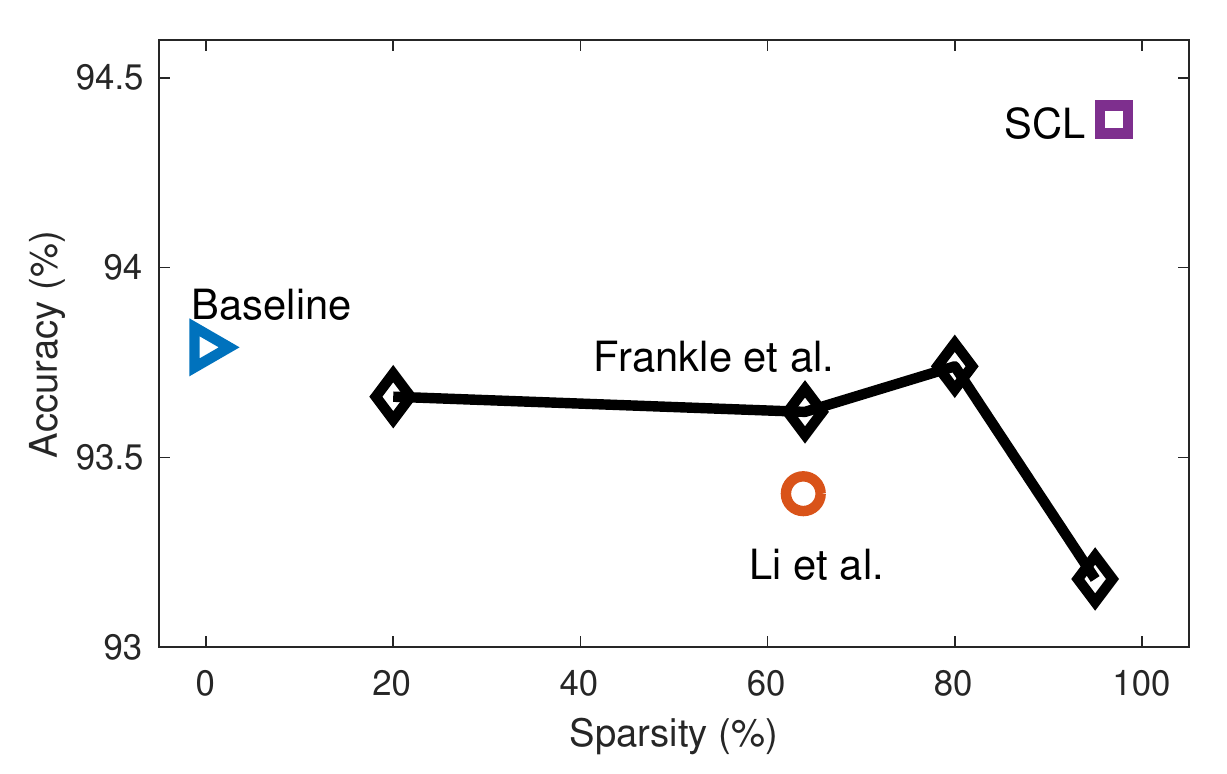}\label{fig:compare.vgg16}}
\hspace{0.3cm}%
\subfigure[VGG-19, comparisons with Han \etal \cite{han2015learning} and Liu \etal \cite{liu2017learning}.]{\includegraphics[width=5cm]{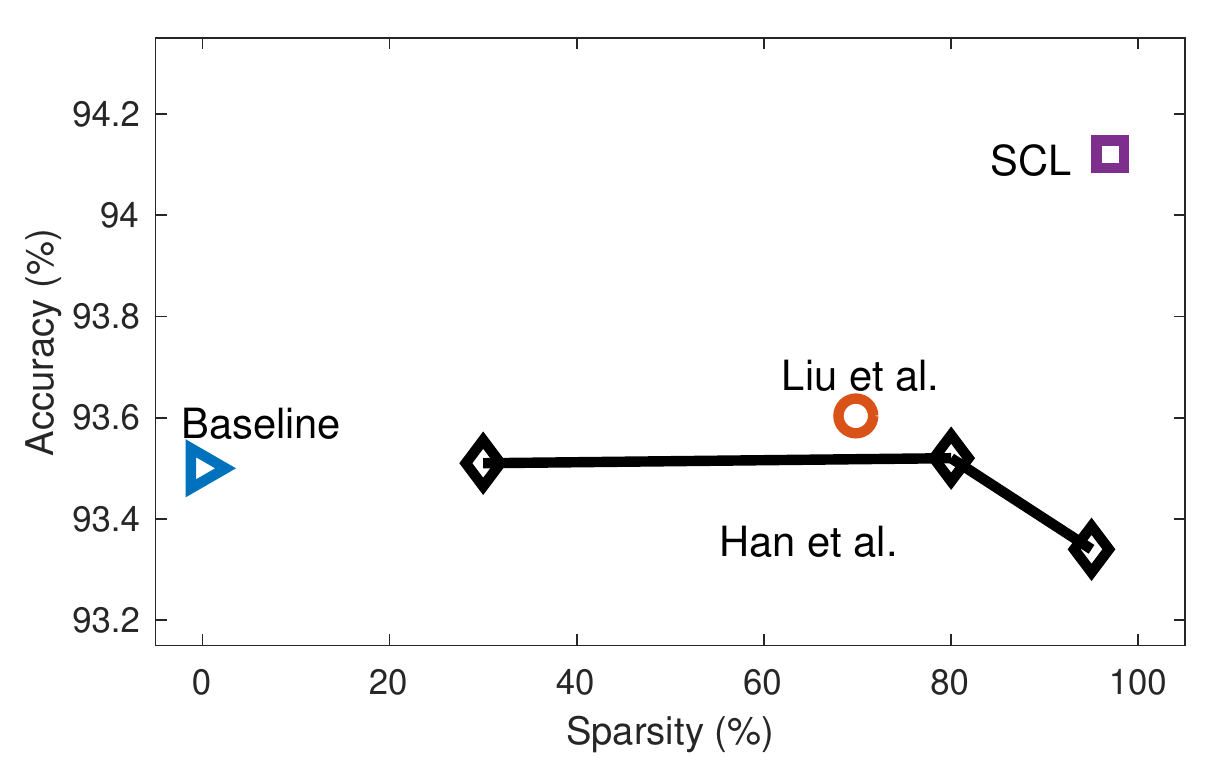}\label{fig:compare:vgg19}}
\hspace{0.3cm}%
\subfigure[DenseNet-BC-100, comparison with Han \etal \cite{han2015learning}.]{\includegraphics[width=5cm]{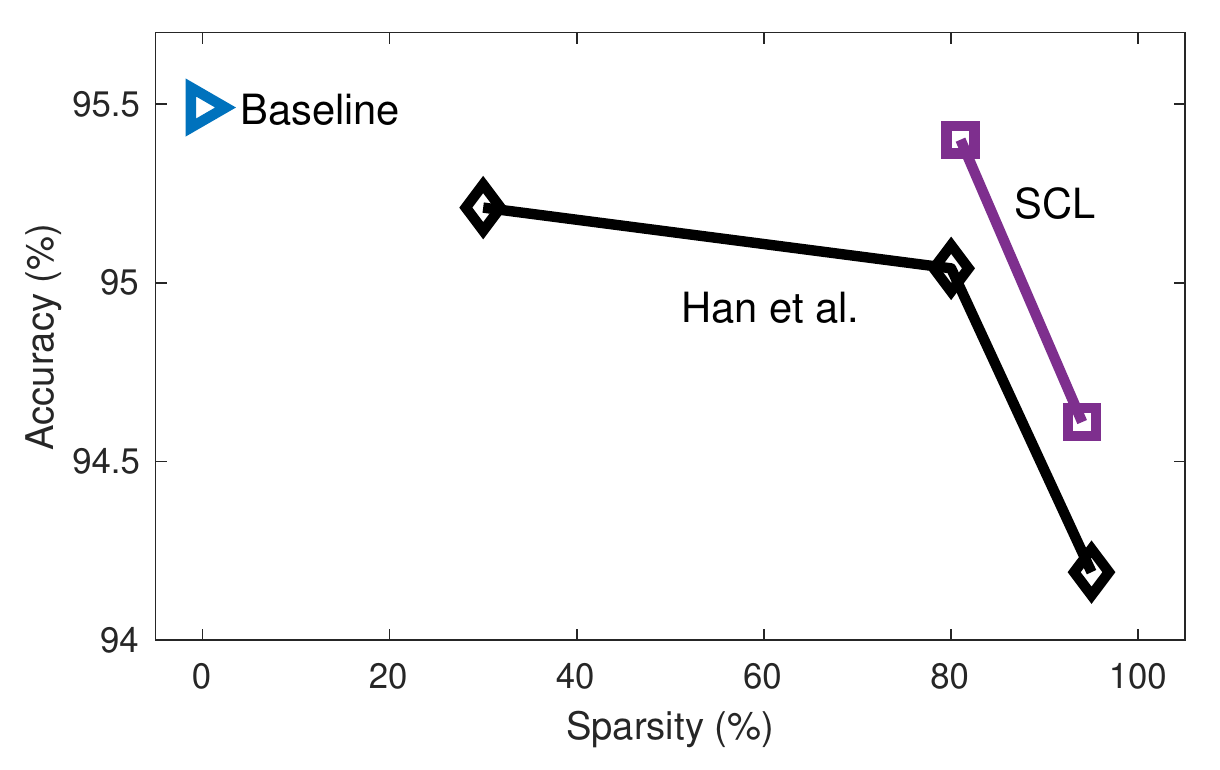}\label{fig:compare:dense-100}}
\caption{Sparsity comparisons with non-structured pruning methods.}
\label{fig:compare_sparsity}
\end{figure*}

\subsection{Comparison with the state-of-the-art pruning methods on CIFAR-10 and CIFAR-100}
\label{sec:sparse_compareCIFAR}
In order to comprehensively evaluate the proposed SCL method, we compare with the state-of-the-art pruning methods in the literature, including both non-structured and structured methods on the baseline networks of VGGs, ResNets, and DenseNets.

\subsubsection{Comparison with non-structured pruning methods} 

We compare the proposed SCL with non-structured pruning methods, \eg, Frankle \etal \cite{frankle2018lottery} and Han \etal \cite{han2015learning}. The results of Frankle \etal \cite{frankle2018lottery} (\ie, Lottery Ticket Hypothesis) are copied from Liu \etal \cite{liu2019rethinking}. As illustrated in Figure \ref{fig:compare_sparsity}, we use a very high sparse target (97\%) to learn sparse VGGs. Our SCL-induced VGGs is superior to the state-of-the-art pruning methods of \cite{frankle2018lottery,han2015learning,li2016pruning,liu2017learning} in both sparsity and accuracy. For DenseNet-BC-100, with the same sparsity of 80\%, our SCL method results in an accuracy of 95.40\%, which is much higher than the accuracy of 95.04\% in Han \etal \cite{han2015learning}. Even though we don't compare FLOPs here due to we are unable to access pre-trained models in these previous works, the reduction in FLOPs is positively correlated with the sparsity for non-structured sparse models. Compared to the human-determined criterion of non-structured pruning methods, the proposed task-aware SCL method yields better results.

\subsubsection{Comparison with structured pruning methods}

As shown in Table \ref{tab:resnet20} and \ref{tab:resnet110}, the ResNet networks trained by our proposed SCL method are more sparse and more accurate. The higher the sparsity, the more savings the FLOPs have. Besides, even with a higher sparsity (\eg, 70\% on ResNet-20 and 90\% on ResNet-110), the SCL-induced ResNet neural networks achieve better accuracy than the baselines in \cite{he2016deep}. In contrast, when the target sparsity exceeds about 30\%, the existing structured pruning methods in \cite{frankle2018lottery} \cite{li2016pruning}\cite{he2018soft}\cite{yu2018nisp} exhibit significant accuracy degradation.

\begin{table}[!t]
  \caption{Results of ResNet-20 on CIFAR-10.}
  \label{tab:resnet20}
  \centering
  \begin{tabular}{@{}
                  c
                  S[table-format=1.3]
                  S[table-format=2]
                  S[table-format=2]
                  S[table-format=2.2]@{}}
    \toprule
     Scheme         & {\# Param.}    &  {Sparsity}  &   {FLOPs $\downarrow$}   & {Accuracy}  \\
    \midrule
Baseline \cite{he2016deep}& 0.268\si{\mega}    &    0\si{\percent}   &  0\si{\percent}  &  91.25\si{\percent}  \\ 
    \midrule
He \etal \cite{he2018soft} & 0.241\si{\mega}        & 10\si{\percent}  &  15\si{\percent}   &  92.24\si{\percent}  \\
He \etal \cite{he2018soft} & 0.214\si{\mega}        & 20\si{\percent}  &  29\si{\percent}   &  91.20\si{\percent}  \\
He \etal \cite{he2018soft} & 0.188\si{\mega}        & 30\si{\percent}  &  42\si{\percent}   &  90.83\si{\percent}  \\
    \midrule
SCL    & 0.133\si{\mega}       & 50\si{\percent}  &   49\si{\percent}  &  92.61\si{\percent}  \\  
SCL    & 0.080\si{\mega}       & 70\si{\percent}  &   69\si{\percent}  &  92.35\si{\percent}  \\  
    \bottomrule
  \end{tabular}
\end{table}

\begin{table}[!t]
  \caption{Results of ResNet-110 on CIFAR-10.}
  \label{tab:resnet110}
  \centering
  \begin{tabular}{@{}
                  c
                  S[table-format=1.2]
                  S[table-format=2]
                  S[table-format=2]
                  S[table-format=2.2]@{}}
    \toprule
     Scheme               & {\# Param.}                  &  {Sparsity}  & {FLOPs $\downarrow$} & {Accuracy} \\
    \midrule
Baseline \cite{he2016deep}& 1.72\si{\mega}                        &    0\si{\percent}  &   0\si{\percent}    &  93.57\si{\percent} \\ 
    \midrule
He \etal \cite{he2018soft}              & 1.20\si{\mega}           & 30\si{\percent}   &  41\si{\percent} &  93.86\si{\percent}  \\
Yu \etal \cite{yu2018nisp}              & 0.98\si{\mega}           & 43\si{\percent}   &  44\si{\percent} &  93.39\si{\percent}  \\
Yu \etal \cite{yu2018nisp}    & 1.17\si{\mega}       & 32\si{\percent}   &   39\si{\percent}    &  93.34\si{\percent}  \\
Li \etal \cite{li2016pruning}  & 1.17\si{\mega}       & 32\si{\percent}   &  39\si{\percent}    &  93.36\si{\percent}  \\
Frankle \etal \cite{frankle2018lottery} & 1.17\si{\mega} & 32\si{\percent} &  39\si{\percent}   &  93.15\si{\percent}  \\
    \midrule
SCL                & 0.51\si{\mega}           & 70\si{\percent}   &    73\si{\percent}     &  94.82\si{\percent} \\ 
SCL                & 0.17\si{\mega}           & 90\si{\percent}   &  90\si{\percent}    &  94.56\si{\percent} \\ 
    \bottomrule
  \end{tabular}
\end{table}

Table \ref{tab:densenet40} lists the experimental results for DenseNet neural networks on the CIFAR-10 dataset. When the target sparsity is moderate (\eg, sparsity $=$ 40\% in Table \ref{tab:densenet40}), our SCL-induced networks show higher accuracy than the baseline in \cite{huang2017densely}. When setting a higher target sparsity (\eg, 65\%), the accuracy of Liu \etal \cite{liu2019rethinking} drops significantly. In contrast, even at a high sparsity of 90\%, our SCL-induced DenseNet network demonstrates little loss of accuracy. The superiority of our SCL is not only due to the advantage of the task-aware feature, but also because the resultant weight-level sparse models have a larger representative capacity.
\begin{table}[!t]
  \caption{Results of DenseNet-40 on CIFAR-10. Results of previous method are copied from Liu \etal \cite{liu2019rethinking}.}
  \label{tab:densenet40}
  \centering
  \begin{tabular}{@{}
                  c
                  S[table-format=1.2]
                  S[table-format=2]
                  S[table-format=2]
                  S[table-format=2.2]@{}}
    \toprule
     Scheme               & {\# Param.}                  &  {Sparsity}  & {FLOPs $\downarrow$} & {Accuracy} \\
    \midrule
Baseline \cite{huang2017densely}        & 1.04\si{\mega}           &    0\si{\percent}  &   0\si{\percent}   &  94.76\si{\percent}  \\ 
    \midrule
Liu \etal \cite{liu2017learning}        & 0.66\si{\mega}           & 36\si{\percent}   &   28\si{\percent}    &  94.81\si{\percent}  \\
Liu \etal \cite{liu2017learning}        & 0.35\si{\mega}           & 65\si{\percent}   &   55\si{\percent}     &  94.35\si{\percent}  \\
    \midrule
SCL                 & 0.62\si{\mega}           & 40\si{\percent}    &   38\si{\percent}   &  94.81\si{\percent}  \\ 
SCL                 & 0.30\si{\mega}           & 71\si{\percent}    &   70\si{\percent}   &  94.66\si{\percent}  \\ 
SCL                 & 0.10\si{\mega}           & 90\si{\percent}    &   88\si{\percent}   &  94.53\si{\percent}  \\ 
    \bottomrule
  \end{tabular}
\end{table}

\subsubsection{Comparison with $L_{0}$ regularization method}
\label{sec:l0}

$L_{0}$ regularization (Gumbel-Softmax trick) \cite{louizos2017learning} determines zero-weight network connections by including a set of non-negative stochastic gates. Compared with \cite{louizos2017learning}, the SCL method has significant differences in the objective function, mask training method, actual reduction of sparse connectivity and FLOPs, as described below. First, the the objective function is different. In this work, the sparse regularization term in the objective function is the $L_{0}$-norm of masks. In contrast, the regularization term in the objective function of \cite{louizos2017learning} is a statistically expected $L_{0}$-norm of masks. Second, the mask training method is different. Stochastic sampling is used in \cite{louizos2017learning} to train mask variables, whereas this work directly trains mask variables without using stochastic sampling. Due to the use of stochastic sampling, the pruning method in \cite{louizos2017learning} is complicated. In contrast, in this work, masks are constrained to 0 or 1 by applying a unit step function on it, and sparsity is produced by penalizing the $L_{0}$-norm of weights. Hence, this work is simple and efficient. Third, the actual reduction of sparse connectivity and FLOPs is different. Both training and testing stages in this work are deterministic, whereas training and testing stages in \cite{louizos2017learning} are stochastic and deterministic, respectively. Note a mask variable implies no connectivity only when its probability is zero. In this work, the probability of connectivity is either 0\% or 100\%, a reduction of $L_{0}$-norm of masks (\ie, probability of certain connectivity is regulated from 100\% to 0\%) in the training stage indicates the same reduction in the testing stage. In contrast, the probability of connectivity is continuous from 0\% to 100\% in the training stage of \cite{louizos2017learning}. Hence, in \cite{louizos2017learning}, the reduction of expected $L_{0}$-norm during training does not necessarily mean a reduction of $L_{0}$-norm during testing. Besides, the discrete mask function is used in \cite{louizos2017learning} for hard pruning in the testing stage. Discrete mask function potentially has a large discrepancy from the probabilistic continuous mask function during training time \cite{kang2020operation}. Furthermore, the use of stochastic sampling in \cite{louizos2017learning} leads to a huge gap between the expected $L_{0}$ in the stochastic training phase and the actual $L_{0}$ in the deterministic testing phase. In \cite{louizos2017learning}, although the hard concrete distribution trick allows zero gates to be produced, the expected $L_{0}$ can not reflect the actual $L_{0}$ during inference.

Our proposed SCL takes advantage of STE \cite{bengio2013estimating} to redefine gradients of mask variables. Even though STE involves the use of STE in stochastic neurons, we think STE is also applicable to deterministic neurons, because deterministic neurons are regarded as a special case of stochastic neurons with a probability of either 0\% (\ie, no connection) or 100\% (\ie, with connection). The use of STE on deterministic problems has been empirically verified by the deep learning community \cite{yin2019understanding,xiao2019autoprune}. Then, $L_{0}$ norm of weights is integrated as a regularization term of the objective function. Despite the role of SCL in training is similar to that of $L_{0}$ regularization, we think SCL is advantageous because its efficient processing of mask variables helps sparsity training. As a result, the proposed SCL method is more direct and effective in encouraging sparse network connections. Table \ref{tab:l0_wrn} lists the experimental results of $L_{0}$ regularization \cite{louizos2017learning} and SCL on the CIFAR-10 and CIFAR-100 datasets. The observation that both of them outperform the baseline model \cite{zagoruyko2016wide} indicates that the best network connections should be sparse. Moreover, the experimental results of SCL with three sparsity levels (\textit{i.e.}, 20\%, 50\%, and 90\%) are provided. Their corresponding accuracy results are better than those of $L_{0}$ regularization, even though the sparsity results are not reported in \cite{louizos2017learning}. In fact, when we use the hyper-parameters and codes in \cite{louizos2017learning} to repeat the experiment for Table \ref{tab:l0_wrn}, we find that the obtained network is not sparse. Although the expected $L_{0}$ norm of weights has been significantly reduced, it is still not low enough to generate sparse connections.

\begin{table}[!t]
  \caption{Comparison with $L_{0}$ regularization \cite{louizos2017learning} on CIFAR datasets. WRN-28-10 \cite{zagoruyko2016wide} is used as the baseline. "-" indicates results not reported.}
  \label{tab:l0_wrn}
  \centering
  \begin{tabular}{@{}
                  c
                  S[table-format=2]
                  S[table-format=2.2]
                  S[table-format=2]
                  S[table-format=2.2]@{}}
    \toprule
\multirow{2}*{Scheme} & \multicolumn{2}{c}{CIFAR-10} & \multicolumn{2}{c}{CIFAR-100} \\
 \cline{2-5}
 & {Sparsity}  &  {Accuracy}  &  {Sparsity}  &  {Accuracy}  \\
 \midrule
Baseline \cite{zagoruyko2016wide} &  0\si{\percent} & 96.00\si{\percent}  & 0\si{\percent}  & 80.75\si{\percent}  \\
    \midrule
Louizos \etal \cite{louizos2017learning}&  {~~-} &96.07\si{\percent}& {~~-}   &  80.96\si{\percent} \\
Louizos \etal \cite{louizos2017learning}& {~~-} & 96.17\si{\percent} &  {~~-} & 81.25\si{\percent} \\
    \midrule
SCL   & 20\si{\percent}  & 96.36\si{\percent}   & 20\si{\percent}  &  81.79\si{\percent} \\ 
SCL   &  51\si{\percent}  &  96.53\si{\percent}     &   50\si{\percent}   & 81.87\si{\percent} \\ 
SCL   & 91\si{\percent}   & 96.33\si{\percent}  &  90\si{\percent} & 81.51\si{\percent}  \\ 
    \bottomrule
  \end{tabular}
\end{table}

\subsubsection{Comparison with existing STE-based pruning method}
\label{sec:l4}
In addition to evaluating the effect of STE and mask gradient normalization in a mapping experiment in Section \ref{sec:experiments_mapping}, we also conduct experiments with DenseNet-based networks on MNIST and with ResNet-20 and ResNet-110 on CIFAR-10, respectively. We compare the results of SCL with the existing STE-based pruning method \cite{xiao2019autoprune}, which uses Leaky ReLU or Softplus STEs without mask gradient normalization. As shown in Tables \ref{tab:mnist_stes} and \ref{tab:cifar_stes}, under the same 95\% sparsity, the result of SCL (\ie, 90.08\%) is superior to \cite{xiao2019autoprune} (\ie, 88.01\% or 87.95\%). Table \ref{tab:cifar_stes} also provides the comparison results of VGG-16 on CIFAR-10, as reported in \cite{xiao2019autoprune}.

\begin{table}[!ht]
  \caption{Comparison Results of DenseNet-based networks on MNIST.}
  \label{tab:mnist_stes}
  \centering
  \begin{tabular}{c|c|lll}
    \toprule
{}  & {Sparsity} & {Leaky ReLU} & {Softplus}   &  {Identity}  \\
      \midrule
{w/o norm}&\multirow{2}{*}{95.0\%}&98.26\% \cite{xiao2019autoprune} &98.31\% \cite{xiao2019autoprune} &98.37\% \\
{w/~ norm}& &98.33\%&98.35\%&98.39\% (SCL) \\
    \bottomrule
  \end{tabular}
\end{table}

\begin{table}[!ht]
  \caption{Comparison of ResNet-20, ResNet-110, and VGG-16 on CIFAR-10.}
  \label{tab:cifar_stes}
  \centering
  \setlength{\tabcolsep}{0.7mm}{
  \begin{tabular}{c|c|c|l|l|l}
    \toprule
 \multicolumn{2}{c|}{}  & {Sparsity} & {Leaky ReLU} & {Softplus}   &  {Identity}  \\
      \midrule
\multirow{2}{*}{ResNet-20}&{w/o norm}&\multirow{2}{*}{95.0\%}&88.01\% \cite{xiao2019autoprune}&87.95\% \cite{xiao2019autoprune}&88.10\% \\
&{w/~ norm}& &89.48\%&90.13\%&90.08\% (SCL) \\
      \midrule
\multirow{2}{*}{ResNet-110}&{w/o norm}&\multirow{2}{*}{95.0\%}&93.15\% \cite{xiao2019autoprune}&92.26\% \cite{xiao2019autoprune}&92.76\% \\
&{w/~ norm}& &93.25\%&93.44\%&93.58\% (SCL)\\
      \midrule
\multirow{2}{*}{VGG-16}&{w/o norm}&98.6\%&\multicolumn{2}{c|}{92.18\% \cite{xiao2019autoprune}} &  \\
&{w/~ norm}& 97.0\%& \multicolumn{2}{c|}{} &94.39\% (SCL)\\
    \bottomrule
  \end{tabular}
  }
\end{table}

\subsection{Comparison with state-of-the-art pruning methods on ImageNet}
\label{sec:sparse_compareIMAGE}
In Tables \ref{tab:imagenet_vgg} and \ref{tab:imagenet_resnet}, except for Han \etal \cite{han2015learning}, the performance results of existing pruning methods on ImageNet are copied from Lin \etal \cite{lin2019toward}. 

In most convolutional neural networks, most of the weights are in fully connected layers, while most FLOPs are in convolutional layers. The number of fully connected layers for VGG-16 and ResNet is three and one, respectively. As shown in Table \ref{tab:imagenet_vgg}, the structured pruning methods of \cite{luo2017thinet,hu2016network,lin2019toward} on VGG-16 lead to low sparsity and significant computational cost reduction, because they only prune the convolutional layers for fewer FLOPs and do not prune fully connected layers. On the other hand, the non-structured pruning methods, \ie, Han \etal~\cite{han2015learning} and our proposed SCL in this study, can obtain higher weight compression ratios, mainly due to the pruning of fully connected layers. Specifically, the method of \cite{han2015learning} achieves a weight sparsity of 92.5\% over its baseline. At a sparsity level of around 90\%, even though the performance (\ie, TOP-1 and TOP-5 drop) of our proposed SCL is worse than \cite{han2015learning}, our SCL method can achieve a much higher accuracy of 69.20\% over 68.66\% in \cite{han2015learning} and a more FLOPs reduction of 87\% over 79\% in \cite{han2015learning}. The difference between our proposed SCL and \cite{han2015learning} is as followed. The pruning efforts in \cite{han2015learning} are mainly for fully connected layers, but less for pruning convolutional layers. In contrast, depending on the weight significance, the proposed SCL can efficiently perform pruning on both fully connected layers and convolutional layers. We also observe that the baseline of \cite{han2015learning} may not converge, because its baseline TOP-1 accuracy is reported as 68.5\%, and our TOP-1 accuracy is larger than 71.5\%. The performance gain in \cite{han2015learning} is attributed to the use of hundreds of training epochs during pruning, which leads to a much better convergence. 

As shown in Table \ref{tab:imagenet_resnet}, for low-sparisty pruning, these cutting-edge pruning methods \cite{li2016pruning,luo2017thinet,wen2016learning,hu2016network,lin2019toward,huang2018data} show significant performance degradation (\ie, TOP-1, TOP-5) than our SCL-induced results. Furthermore, our SCL method achieves a high sparsity of 74\% with a FLOPs reduction of 79\%, while retaining a slight performance degradation.  

Compared with \cite {he2020learning}, this work has three main differences. First, our SCL method is pruning criterion-free, whereas \cite {he2020learning} intends to learn proper layer-wise pruning criterion from a set of designer-defined criteria. Due to the criterion-free nature, our SCL method does not need hyper-parameters in criterion. In contrast, \cite {he2020learning} needs hyper-parameters to determine the number of filters to keep. Second, the pruning performance of \cite {he2020learning} heavily depends on human experience, including designer-defined hyper-parameters and criteria set. In contrast, our SCL method automatically learns the optimized network connectivity in a task-aware manner. Third, binary masks and weight parameters are jointly updated in SCL, whereas mask and weight parameters are trained separately in \cite {he2020learning}. As shown in Table \ref{tab:imagenet_resnet}, this work shows better network pruning results than \cite {he2020learning}. SCL achieves a much lower drop of TOP-1 accuracy (\ie, 0.23\%) than \cite {he2020learning} (\ie, 1.69\%) and a much lower drop of TOP-5 accuracy (\ie, 0.40\%) than \cite {he2020learning} (\ie, 0.83\%), meanwhile largely reducing the number of FLOPs (\ie, 79\%) than \cite {he2020learning} (\ie, 61\%).

Compared with \cite{huang2018data}, this work has three main differences. First, our SCL method uses binary masks to represent network connectivity, whereas \cite{huang2018data} uses continuous scaling factors to represent network connectivity. In the proposed SCL method, no threshold is needed to train binary masks. In contrast, soft-threshold needs to be trained to obtain non-negative scaling factors in \cite{huang2018data}. In \cite{huang2018data}, if the value of a scaling factor is zero, it indicates no network connection. In \cite{huang2018data}, a positive scaling factor indicates the existence of network connection. Second, our SCL method uses scheduled SGD to train the binary masks, whereas \cite{huang2018data} uses APG to train the scaling factor parameters. Third, \cite{huang2018data} only prunes convolutional layers to reduce FLOPs, while our SCL method prunes convolutional layers and fully-connected layers to compress the weight size. As a result, the SCL method leads to much higher sparsity in network connectivity, and therefore fewer trainable parameters. As shown in Table \ref{tab:imagenet_resnet}, this work shows better network pruning performance than \cite{huang2018data}. SCL achieves a much lower drop of TOP-1 accuracy (\ie, 0.23\%) and a much lower drop of TOP-5 accuracy (\ie, 0.40\%) than \cite{huang2018data}, meanwhile reducing the number of parameters by 58\% (\ie, 6.6M vs. 15.6M).

Even though this work and \cite {xiao2019autoprune} are all inspired by the general concept of STE, the objective function optimization, update rule for mask parameters, and coarse gradient  estimation used in this work are significantly different from \cite {xiao2019autoprune}. In this work, the weight and mask parameters are updated in the same optimization iteration. Yet, the weight and mask parameters are updated separately in \cite {xiao2019autoprune}, which corresponds to high computational complexity. As a result, the training calculation cost of SCL is significantly reduced. In this work, the update rule is not based on assumptions, and the update rule for mask parameters is derived through the back-propagation algorithm. In contrast, gradients in \cite {xiao2019autoprune} are modified by dividing true mask gradients (\ie, gradients obtained through back-propagation) by the absolute value of weights element-wisely. Two coarse gradient estimations (\ie, gradients of Leaky ReLU and Softplus function) are used in \cite {xiao2019autoprune}, whereas the straight-through gradient estimation (\ie, gradient of identity function) is used in this work.  As shown in Table \ref{tab:imagenet_resnet}, in addition to the significant reduction of FLOPs (\ie, 79\% in this work vs. 55\% in \cite {xiao2019autoprune}), our SCL pruning method achieves a much lower drop of TOP-1 accuracy (\ie, 0.23\% in this work vs. 0.40\% in \cite {xiao2019autoprune}). These experimental results demonstrate that our SCL method is better than \cite {xiao2019autoprune}.

Compared with \cite{kang2020operation}, this work has two main differences. First, the SCP method in \cite{kang2020operation} assumes that feature maps follow a Gaussian distribution.  Yet, this assumption is too strict to derive accurate gradients. In contrast, our SCL method does not rely on any assumptions. Our SCL method  trains network connectivity through STE gradient estimation and gradient back-propagation. Second, the SCP method in \cite{kang2020operation} has a good pruning performance in network layers that are followed by BN and ReLU. However, there are no BN and ReLU in many deep neural networks. For example, BN does not exist in the VGGs network. In the architectures of MobileNets or EfficientNets, some convolutional layers are only followed by BN rather than BN and ReLU. As shown in Table 5 of \cite{kang2020operation}, the network pruning performance is significantly degraded if only BN exists for channel pruning. As a result, the SCP method in \cite{kang2020operation} does not show good pruning performance for many neural networks. In contrast, experimental results in this work show that our SCL method is applicable to various neural networks, including VGGs, DenseNets, ResNets, EfficientNets, and RNNs. As shown in Table \ref{tab:imagenet_resnet}, SCL shows better network pruning performance than \cite{kang2020operation}. SCL achieves a much lower drop of TOP-1 accuracy (\ie, 0.23\%) than \cite{kang2020operation} (\ie, 1.69\%) and a much lower drop of TOP-5 accuracy (\ie, 0.40\%) than \cite{kang2020operation} (\ie, 0.98\%), meanwhile largely reducing the number of FLOPs (\ie, 79\%) than \cite{kang2020operation} (\ie, 54\%). 

We also apply SCL to the EfficientNet \cite{tan2019efficientnet} architecture to learn sparse connections. Unlike VGG, ResNet, and DenseNet architectures that are designed by humans, the EfficientNet architecture is automatically determined by a neural architecture search technique. Depth-wise convolution is widely used in EfficientNet architectures to improve efficiency. As shown in Table \ref{tab:effcientnet}, three pruning polices are used by us for depth-wise convolution in the experiments of \cite{zhu2017prune} on the ImageNet dataset. In the first pruning policy, the pruning rates of the convolution, depth-wise convolution, and classifier layers are set to 0.2, 0.1, and 0.2, respectively. In the second pruning policy, the pruning rates of the convolution, depth-wise convolution, and classifier layers are set to 0.5, 0.5, and 0.5, respectively. In the third pruning policy, the pruning rates of the convolution, depth-wise convolution, and classifier layers are set to 0.5, 0.125, and 0.6, respectively. Compared with the baseline EfficientNet-B0, Table \ref{tab:effcientnet} demonstrates that when the sparsity is low (such as 19.5\% and 25.8\%), both the pruning method of Zhu \etal \cite{zhu2017prune} and SCL show a negligible accuracy degradation. According to the magnitude of weights, the work of Zhu \etal \cite{zhu2017prune} forces smaller weights to zero. As a result, although it is an element-wise pruning method, if the difference of weight magnitudes between depth-wise channels is large, the pruning method of Zhu \etal \cite{zhu2017prune} tends to completely remove certain depth-wise convolution channels. The second policy of Zhu \etal \cite{zhu2017prune} sets the weights of all layers to be pruned by 50\%, experimental results show that FLOPs are reduced by 60\% with a big TOP-1 accuracy drop of 0.79\%. The third policy of Zhu \etal \cite{zhu2017prune} prunes less weights of depth-wise layers and obtains a better accuracy. These pruning results of Zhu \etal \cite{zhu2017prune} show that it is difficult to find an appropriate pruning policy to obtain satisfactory pruning results. In contrast, since the network connectivity learned by SCL is determined by the significance of weight, SCL automatically retains necessary depth-wise channels even if their magnitudes of weight are small. As a result, when SCL prunes more weights than the pruning method of Zhu \etal \cite{zhu2017prune} (\textit{i.e.}, a sparsity of 55.0\% versus 50.9\%), SCL significantly improves the accuracy results. Thus, SCL also outperforms the pruning method of Zhu \etal \cite{zhu2017prune} in terms of EfficientNet architectures.

\begin{table*}[!t]
  \caption{Pruning results of VGG-16 on ImageNet.}
  \label{tab:imagenet_vgg}
  \centering
  \begin{tabular}{@{}
                  c
                  S[table-format=3.1]
                  c
                  S[table-format=2]
                  l
                  l@{}}
    \toprule
     Scheme         & {\# Param.}    &  {Sparsity}  &   {FLOPs $\downarrow$}   & {TOP-1 $\downarrow$ (TOP-1)} & {TOP-5 $\downarrow$ (TOP-5)}\\
    \midrule  
Li \etal \cite{li2016pruning} &  126.7\si{\mega}        & 8.38\si{\percent}  &  71\si{\percent}   &  ~0.29\si{\percent}  &  -0.05\si{\percent}\\
Luo \etal \cite{luo2017thinet} & 131.5\si{\mega}        & 4.92\si{\percent}  &  68\si{\percent}   &  -1.46\si{\percent}  &  -1.09\si{\percent}\\
Hu \etal \cite{hu2016network} &  126.7\si{\mega}        & 8.38\si{\percent}  & 71\si{\percent}   &  -0.64\si{\percent}  &  -0.43\si{\percent}\\
Lin \etal \cite{lin2019toward} & 126.2\si{\mega}        & 8.75\si{\percent}  &  71\si{\percent}  &  -1.65\si{\percent}  &  -0.97\si{\percent}  \\
Han \etal \cite{han2015learning} & 10.3\si{\mega}     & 92.5\si{\percent}  &  79\si{\percent}   &  -0.26\si{\percent} (68.66\si{\percent}) &  ~0.44\si{\percent} (89.12\si{\percent}) \\
    \midrule
SCL    & 60.2\si{\mega}       & 56.5\si{\percent}  &  50\si{\percent}   &  -1.11\si{\percent} (72.84\si{\percent}) &  -0.54\si{\percent} (90.88\si{\percent}) \\ 
SCL    & 36.9\si{\mega}       & 73.3\si{\percent}  &  71\si{\percent}  &  -0.33\si{\percent} (72.05\si{\percent})  &  -0.25\si{\percent} (90.60\si{\percent}) \\ 
SCL    & 13.9\si{\mega}       & 89.9\si{\percent}  &  87\si{\percent}   &  ~2.49\si{\percent} (69.24\si{\percent})  &  ~1.11\si{\percent} (89.24\si{\percent}) \\ 
    \bottomrule
  \end{tabular}
\end{table*}

\begin{table}[!t]
  \caption{Pruning results of ResNet-50 on ImageNet.}
  \label{tab:imagenet_resnet}
  \centering
  \setlength{\tabcolsep}{1mm}{
  \begin{tabular}{@{}
                  c
                  S[table-format=2.1]
                  S[table-format=2]
                  S[table-format=2]
                  S[table-format=1.2]
                  S[table-format=1.2]@{}}
    \toprule
     Scheme         & {\# Param.}    &  {Sparsity}  &   {FLOPs $\downarrow$}   & {TOP-1 $\downarrow$} & {TOP-5 $\downarrow$} \\
    \midrule   
Li \etal \cite{li2016pruning} & 15.9\si{\mega}     & 38\si{\percent}  &  54\si{\percent}   &  3.36\si{\percent}  &  2.08\si{\percent}\\
Li \etal \cite{li2016pruning} & 12.2\si{\mega}     & 52\si{\percent}  &  59\si{\percent}   &  4.31\si{\percent}  &  2.42\si{\percent}\\
Luo \etal \cite{luo2017thinet} & 16.9\si{\mega}    & 34\si{\percent}  &  41\si{\percent}    &  3.09\si{\percent}  &  1.63\si{\percent}\\
Luo \etal \cite{luo2017thinet} & 12.3\si{\mega}    & 52\si{\percent}  &  59\si{\percent}    &  4.12\si{\percent}  &  2.28\si{\percent}\\
Wen \etal \cite{wen2016learning} & 13.2\si{\mega}  & 48\si{\percent}  &  49\si{\percent}    &  4.58\si{\percent}  &  2.68\si{\percent}\\
Hu \etal \cite{hu2016network} & 15.9\si{\mega}     & 38\si{\percent}  &  54\si{\percent}    &  3.47\si{\percent}  &  2.39\si{\percent}\\
Hu \etal \cite{hu2016network} & 12.2\si{\mega}     & 52\si{\percent}  &  59\si{\percent}    &  4.25\si{\percent}  &  2.41\si{\percent}\\
Lin \etal \cite{lin2019toward} & 15.5\si{\mega}   & 39\si{\percent}  &  54\si{\percent}  &  2.83\si{\percent}  &  1.57\si{\percent}  \\
Lin \etal \cite{lin2019toward} & 12.0\si{\mega}   & 53\si{\percent}  &  59\si{\percent}  &  3.65\si{\percent}  &  2.11\si{\percent}  \\
He \etal \cite{he2020learning} & {~~-}     & {~~-}  &  61\si{\percent}    &  1.69\si{\percent}  &  0.83\si{\percent}\\
Huang \etal \cite{huang2018data} & 15.6\si{\mega}     & 39\si{\percent}  &  43\si{\percent}    &  4.30\si{\percent}  &  2.07\si{\percent}\\
Xiao \etal \cite{xiao2019autoprune} & {~~-}     & {~~-}  &  55\si{\percent}    &  0.40\si{\percent}  &  {~~-}\\
Kang \etal \cite{kang2020operation} & {~~-}     & {~~-}  &  54\si{\percent}    &  1.69\si{\percent}  &  0.98\si{\percent}\\
    \midrule
SCL    & 17.9\si{\mega}       & 30\si{\percent}  &   24\si{\percent}   &  -0.30\si{\percent}   &  -0.16\si{\percent} \\  
SCL    & 6.6\si{\mega}        & 74\si{\percent}  &  79\si{\percent}   &  0.23\si{\percent}  &  0.40\si{\percent}\\ 
    \bottomrule
  \end{tabular}
  }
\end{table}

\begin{table}[!t]
  \caption{Pruning results of EfficientNet-b0 \cite{tan2019efficientnet} on ImageNet. $^1$, $^2$, and $^3$ indicate the first, second, and third pruning policy, respectively.
  }
  \label{tab:effcientnet}
  \centering
  \setlength{\tabcolsep}{1mm}{
  \begin{tabular}{@{}
                  c
                  S[table-format=1.2]
                  S[table-format=2.1]
                  S[table-format=2]
                  S[table-format=1.2]
                  S[table-format=1.2]@{}}
    \toprule
     Scheme   & {\# Param.}  &  {Sparsity} &  {FLOPs $\downarrow$}  &    {TOP-1 $\downarrow$}      &  {TOP-5 $\downarrow$}\\
    \midrule
Zhu \etal \cite{zhu2017prune}$^1$ &  4.24\si{\mega}      &   19.5\si{\percent}    & 18\si{\percent} &  0.04\si{\percent}  & -0.04\si{\percent}\\  
Zhu \etal \cite{zhu2017prune}$^2$&  2.65\si{\mega}      &   49.7\si{\percent} & 60\si{\percent} &  0.79\si{\percent}      & 0.26\si{\percent}\\ 
Zhu \etal \cite{zhu2017prune}$^3$&  2.59\si{\mega}     &   50.9\si{\percent}   & 51\si{\percent} &  0.41\si{\percent}      & 0.18\si{\percent} \\ 


    \midrule
SCL                       &  3.91\si{\mega}     &   25.8\si{\percent}     & 24\si{\percent} & 0.02\si{\percent}      & 0.01\si{\percent}\\
SCL                       &  2.37\si{\mega}      &   55.0\si{\percent}     & 53\si{\percent} & 0.32\si{\percent}      & 0.13\si{\percent}\\
    \bottomrule
  \end{tabular}
  }
\end{table}

\subsection{Sparse RNN learning on WikiText-2}

When SCL prunes weight matrices in RNNs (\ie, setting some weight values to zero), it does not necessarily produce zero gradients. In addition, the network connectivity of RNNs is not sparsely initialized in SCL. During the training process, sparse connections are gradually produced by SCL. Hence, SCL does not cause the exploding/vanishing gradient problem for RNNs. In this work,
SCL is applied to a word-level language model WikiText-2 \cite{merity2017pointer} for verifying its effectiveness on RNNs. The perplexity of the language model is evaluated. The lower the perplexity, the better the RNN model. Table \ref{tab:rnn} lists the experimental results of existing pruning methods (\textit{i.e.}, Zhu \cite{zhu2017prune} and Narang \cite{narang2017exploring}) and SCL. To obtain these results, the pruning method of Zhu \etal \cite{zhu2017prune} has to find an appropriate combination of sparse rates for each weight tensor through a lot of trials and errors, while the sparsity for each weight is automatically found by SCL. Compared with the baseline model that has a sparsity of 0\%, sparse models obtained by these existing state-of-the-art pruning methods and SCL achieve higher perplexity values. The 80\% sparse RNN model trained by Zhu \etal \cite{zhu2017prune} or SCL reduces the number of weight parameters by almost 80\% and meanwhile outperforms the baseline in terms of perplexity on the test dataset. The results indicate that SCL can effectively output excellent sparse connectivity for RNNs. Besides, when the expected sparsity is 95\%, the perplexity results of SCL are better than those of Zhu \etal \cite{zhu2017prune} and Narang \cite{narang2017exploring}. 

\begin{table}[!t]
  \caption{Results of sparse RNN learning on WikiText-2 \cite{merity2017pointer}. Perplexity of the language models is evaluated, the lower perplexity the better.}
  \label{tab:rnn}
  \centering
  \setlength{\tabcolsep}{1mm}{
  \begin{tabular}{@{}
                  c
                  S[table-format=2.1]
                  S[table-format=2.1]
                  S[table-format=3.2]
                  S[table-format=2.2]@{}}
    \toprule
     Scheme   & {\# Param.}  &  {Sparsity}  &    {Perpl. (Validation)}      &  {Perpl. (Test)}\\
    \midrule
Baseline      &  85.9\si{\mega}     &    0\si{\percent}     &  87.49      & 83.85\\
    \midrule

%
Zhu \etal \cite{zhu2017prune}&  16.8\si{\mega}     &   80.4\si{\percent}   &  89.31      & 83.64\\
Zhu \etal \cite{zhu2017prune}&  8.6\si{\mega}      &   90.0\si{\percent} &  90.70      & 85.67\\
Narang \etal \cite{narang2017exploring}&4.9\si{\mega} &94.3\si{\percent}  &  100.23      & 95.35\\ 
Zhu \etal \cite{zhu2017prune}&  4.3\si{\mega}      &   95.0\si{\percent}    &  98.42  & 92.79\\
    \midrule
SCL                       &  16.1\si{\mega}     &   81.3\si{\percent}     &  88.97      & 83.16\\
SCL                       &  4.6\si{\mega}      &   94.7\si{\percent}     &  97.63      & 91.27\\
    \bottomrule
  \end{tabular}
  }
\end{table}

\subsection{Applicability of SCL in Sparse IndRNN learning }

IndRNN \cite{li2018independently} has recently been proposed to solve the exploding/vanishing gradient problem in RNNs. IndRNN modifies and updates the RNN states from
\begin{equation}
\label{eq:conventional_rnn}
  \textbf{h}_{t} = \sigma (\textbf{W} \textbf{x}_{t} + \textbf{U} \textbf{h}_{t-1} + \textbf{b})
\end{equation}
to
\begin{equation}
\label{eq:ind_rnn}
  \textbf{h}_{t} = \sigma (\textbf{W} \textbf{x}_{t} + \textbf{u} \odot \textbf{h}_{t-1} + \textbf{b})
\end{equation}
where $\textbf{x}_{t} \in \mathbb{R}^{M}$ and $\textbf{h}_{t} \in \mathbb{R}^{N}$ are the input states and hidden states at a time step $t$, respectively. $\textbf{W} \in \mathbb{R}^{N \times M}$, $\textbf{U} \in \mathbb{R}^{N \times N}$, and $\textbf{b} \in \mathbb{R}^{N}$ represent the weights for the current input, recurrent input, and bias of neurons, respectively. $\textbf{u} \in \mathbb{R}^{N}$ is a weight vector. When the weight matrix $\textbf{U}$ happens to be a diagonal matrix, the vector $\textbf{u}$ can be regarded as the diagonal vector of matrix $\textbf{U}$. $N$ represents the number of neurons in this layer and $ \sigma $ represents an activation function.  We can see that $\textbf{U} \textbf{h}_{t-1}$ in the RNN is reformulated as Hadamard product $\textbf{u} \odot \textbf{h}_{t-1}$ in the IndRNN. Therefore, the gradient of the $n$-th neuron at the time step $t$ is changed from the RNN gradient
\begin{equation}
\label{eq:conventional_rnn_gradient}
  \frac{\partial J}{\partial h_{t}} = \frac{\partial J}{\partial h_{T}} \prod _{k=t} ^{T-1} diag( \sigma^{'}(h_{k+1})) \textbf{U}^{T}
\end{equation}
to the IndRNN gradient
\begin{equation}
\label{eq:ind_rnn_gradient}
  \frac{\partial J_{n}}{\partial h_{n,t}} = \frac{\partial J_{n}}{\partial h_{n,T}} u_{n}^{T-t} \prod _{k=t} ^{T-1} \sigma^{'}_{n, k+1}
\end{equation}
where $diag( \sigma^{'}(h_{k+1}))$ is the Jacobian matrix of the
element-wise activation function. Due to the term of $\prod _{k=t} ^{T-1} diag( \sigma^{'}(h_{k+1})) \textbf{U}^{T}$, it is difficult to control the gradient of a RNN within a appropriate range. Fortunately, the exploding/vanishing gradient problem is easily addressed in the IndRNN by regulating the exponential term of $u_{n}^{T-t} \prod _{k=t} ^{T-1} \sigma^{'}_{n, k+1}$ within an appropriate range during the training process.

Next, let us analyze the impact of our SCL method on the exploding/vanishing gradient problem in IndRNN neural networks.
For IndRNN networks, the proposed SCL method is applicable to prune weights in the matrix $\textbf{W}$ in the IndRNN without affecting the exploding/vanishing gradient problem in training. Moreover, the weights in the vector $\textbf{u}$ should be excluded from being pruned by SCL, because if some weights in $\textbf{u}$ are pruned, some values of $u_{n}^{T-t}$ will be set to zero, thus causing the vanishing gradient problem in the IndRNN. In fact, by comparing the dimension of weight matrix $\textbf{W} \in \mathbb{R}^{N \times M}$ and weight vector $\textbf{u} \in \mathbb{R}^{N}$, we see that the majority of weight parameters are located in the matrix $\textbf{W}$. As a result, when SCL prunes the weight matrix $\textbf{U}$ and meanwhile prevents the vector $\textbf{u}$ from being pruned, the pruning ability and space in the IndRNN networks are not significantly reduced. Since we have shown the effectiveness of the proposed SCL on RNNs, both of whose weight matrices $\textbf{W}$ and $\textbf{U}$ are pruned by SCL, it is expected that when only pruning the weight matrix $\textbf{W}$, the proposed SCL method can achieve highly sparse IndRNN networks without affecting the exploding/vanishing gradient problem.

\section{Conclusion}
\label{sec:concusion}

We present a Sparse Connectivity Learning (SCL) method to automatically explore and optimize sparse network connectivity. As the number of neural network connections is incorporated into the objective function, the network connectivity can be optimized for a given sparsity expectation to achieve the best performance. Our proposed SCL method has the task-aware ability, which does not require designer-defined pruning criteria or hyper-parameters for each network layer. As a result, the SCL-induced sparse networks are explored in a larger hypothesis space, and they have the potential to generate optimized network connections. The proposed SCL is applicable to various neural network architectures including fully connected networks, convolutional neural networks (VGGs, ResNets, DenseNets, and EfficientNets), and recurrent neural networks (RNNs). Experiments on the MNIST, CIFAR-10, CIFAR-100, ImageNet, and WikiText-2 datasets demonstrate that the proposed SCL method achieves highly efficient learning of sparse network connectivity in network compression ratio, FLOP reduction, and accuracy over existing state-of-the-art pruning methods in the literature. So far, SCL supports convolutions, fully connected layers, and RNN layers. We will explore SCL on other operators in the future. 

\section*{Acknowledgement}
This work is partially supported by China Natural Science Foundation under grant (No. 62171391), the Opening Foundation of Yulin Research Institute of Big Data (Grant No. 2020YJKY04).

We thank Dr. Chun-Chen Liu from Kneron Inc. who provided insight and expertise that greatly assisted the research. We would also like to acknowledge Cheng-Hung Hsieh, Jia-En Hsieh from National Chiao Tung University for their helpful discussion.








\bibliographystyle{IEEEtran}
\bibliography{IEEEabrv,bib}

\begin{thebibliography}{10}
\providecommand{\url}[1]{#1}
\csname url@samestyle\endcsname
\providecommand{\newblock}{\relax}
\providecommand{\bibinfo}[2]{#2}
\providecommand{\BIBentrySTDinterwordspacing}{\spaceskip=0pt\relax}
\providecommand{\BIBentryALTinterwordstretchfactor}{4}
\providecommand{\BIBentryALTinterwordspacing}{\spaceskip=\fontdimen2\font plus
\BIBentryALTinterwordstretchfactor\fontdimen3\font minus
  \fontdimen4\font\relax}
\providecommand{\BIBforeignlanguage}[2]{{%
\expandafter\ifx\csname l@#1\endcsname\relax
\typeout{** WARNING: IEEEtran.bst: No hyphenation pattern has been}%
\typeout{** loaded for the language `#1'. Using the pattern for}%
\typeout{** the default language instead.}%
\else
\language=\csname l@#1\endcsname
\fi
#2}}
\providecommand{\BIBdecl}{\relax}
\BIBdecl

\bibitem{szegedy2015going}
C.~Szegedy, W.~Liu, Y.~Jia, P.~Sermanet, S.~Reed, D.~Anguelov, D.~Erhan,
  V.~Vanhoucke, and A.~Rabinovich, ``Going deeper with convolutions,'' in
  \emph{Proceedings of the IEEE conference on computer vision and pattern
  recognition}, 2015, pp. 1--9.

\bibitem{krizhevsky2012imagenet}
A.~Krizhevsky, I.~Sutskever, and G.~E. Hinton, ``Imagenet classification with
  deep convolutional neural networks,'' in \emph{Advances in neural information
  processing systems}, 2012, pp. 1097--1105.

\bibitem{simonyan2014very}
K.~Simonyan and A.~Zisserman, ``Very deep convolutional networks for
  large-scale image recognition,'' in \emph{ICLR}, 2015.

\bibitem{he2016deep}
K.~He, X.~Zhang, S.~Ren, and J.~Sun, ``Deep residual learning for image
  recognition,'' in \emph{Proceedings of the IEEE conference on computer vision
  and pattern recognition}, 2016, pp. 770--778.

\bibitem{huang2017densely}
G.~Huang, Z.~Liu, L.~Van Der~Maaten, and K.~Q. Weinberger, ``Densely connected
  convolutional networks,'' in \emph{Proceedings of the IEEE conference on
  computer vision and pattern recognition}, 2017, pp. 4700--4708.

\bibitem{lecun2015deep}
Y.~LeCun, Y.~Bengio, and G.~Hinton, ``Deep learning,'' \emph{nature}, vol. 521,
  no. 7553, p. 436, 2015.

\bibitem{han2015learning}
S.~Han, J.~Pool, J.~Tran, and W.~Dally, ``Learning both weights and connections
  for efficient neural network,'' in \emph{Advances in neural information
  processing systems}, 2015, pp. 1135--1143.

\bibitem{guo2016dynamic}
Y.~Guo, A.~Yao, and Y.~Chen, ``Dynamic network surgery for efficient dnns,'' in
  \emph{Advances In Neural Information Processing Systems}, 2016, pp.
  1379--1387.

\bibitem{ullrich2017soft}
K.~Ullrich, E.~Meeds, and M.~Welling, ``Soft weight-sharing for neural network
  compression,'' in \emph{ICLR}, 2017.

\bibitem{molchanov2017variational}
D.~Molchanov, A.~Ashukha, and D.~Vetrov, ``Variational dropout sparsifies deep
  neural networks,'' in \emph{ICML}, 2017.

\bibitem{zhu2017prune}
M.~Zhu and S.~Gupta, ``To prune, or not to prune: exploring the efficacy of
  pruning for model compression,'' \emph{arXiv preprint arXiv:1710.01878},
  2017.

\bibitem{tartaglione2018learning}
E.~Tartaglione, S.~Leps{\o}y, A.~Fiandrotti, and G.~Francini, ``Learning sparse
  neural networks via sensitivity-driven regularization,'' in \emph{Advances in
  Neural Information Processing Systems}, 2018, pp. 3878--3888.

\bibitem{frankle2018lottery}
J.~Frankle and M.~Carbin, ``The lottery ticket hypothesis: Finding sparse,
  trainable neural networks,'' in \emph{ICLR}, 2019.

\bibitem{li2016pruning}
H.~Li, A.~Kadav, I.~Durdanovic, H.~Samet, and H.~P. Graf, ``Pruning filters for
  efficient convnets,'' in \emph{ICLR}, 2017.

\bibitem{luo2017thinet}
J.-H. Luo, J.~Wu, and W.~Lin, ``Thinet: A filter level pruning method for deep
  neural network compression,'' in \emph{ICCV}, 2017, pp. 5058--5066.

\bibitem{he2018soft}
Y.~He, G.~Kang, X.~Dong, Y.~Fu, and Y.~Yang, ``Soft filter pruning for
  accelerating deep convolutional neural networks,'' \emph{IJCAI}, 2018.

\bibitem{lin2018accelerating}
S.~Lin, R.~Ji, Y.~Li, Y.~Wu, F.~Huang, and B.~Zhang, ``Accelerating
  convolutional networks via global \& dynamic filter pruning.'' in
  \emph{IJCAI}, 2018, pp. 2425--2432.

\bibitem{lin2019toward}
S.~Lin, R.~Ji, Y.~Li, C.~Deng, and X.~Li, ``Toward compact convnets via
  structure-sparsity regularized filter pruning,'' \emph{IEEE transactions on
  neural networks and learning systems}, 2019.

\bibitem{he2019filter}
Y.~He, P.~Liu, Z.~Wang, Z.~Hu, and Y.~Yang, ``Filter pruning via geometric
  median for deep convolutional neural networks acceleration,'' in \emph{CVPR},
  2019.

\bibitem{liu2017learning}
Z.~Liu, J.~Li, Z.~Shen, G.~Huang, S.~Yan, and C.~Zhang, ``Learning efficient
  convolutional networks through network slimming,'' in \emph{ICCV}, 2017, pp.
  2755--2763.

\bibitem{he2017channel}
Y.~He, X.~Zhang, and J.~Sun, ``Channel pruning for accelerating very deep
  neural networks,'' in \emph{ICCV}, 2017.

\bibitem{zhuang2018discrimination}
Z.~Zhuang, M.~Tan, B.~Zhuang, J.~Liu, Y.~Guo, Q.~Wu, J.~Huang, and J.~Zhu,
  ``Discrimination-aware channel pruning for deep neural networks,'' in
  \emph{Advances in Neural Information Processing Systems}, 2018, pp. 875--886.

\bibitem{liu2019channel}
C.~Liu and H.~Wu, ``Channel pruning based on mean gradient for accelerating
  convolutional neural networks,'' \emph{Signal Processing}, vol. 156, pp.
  84--91, 2019.

\bibitem{li2019oicsr}
J.~Li, Q.~Qi, J.~Wang, C.~Ge, Y.~Li, Z.~Yue, and H.~Sun, ``Oicsr:
  Out-in-channel sparsity regularization for compact deep neural networks,'' in
  \emph{CVPR}, 2019.

\bibitem{wen2016learning}
W.~Wen, C.~Wu, Y.~Wang, Y.~Chen, and H.~Li, ``Learning structured sparsity in
  deep neural networks,'' in \emph{Advances in Neural Information Processing
  Systems}, 2016, pp. 2074--2082.

\bibitem{anwar2017structured}
S.~Anwar, K.~Hwang, and W.~Sung, ``Structured pruning of deep convolutional
  neural networks,'' \emph{ACM Journal on Emerging Technologies in Computing
  Systems}, vol.~13, no.~3, p.~32, 2017.

\bibitem{zhang2018shufflenet}
X.~Zhang, X.~Zhou, M.~Lin, and J.~Sun, ``Shufflenet: An extremely efficient
  convolutional neural network for mobile devices,'' in \emph{Proceedings of
  the IEEE Conference on Computer Vision and Pattern Recognition}, 2018, pp.
  6848--6856.

\bibitem{howard2017mobilenets}
A.~G. Howard, M.~Zhu, B.~Chen, D.~Kalenichenko, W.~Wang, T.~Weyand,
  M.~Andreetto, and H.~Adam, ``Mobilenets: Efficient convolutional neural
  networks for mobile vision applications,'' \emph{arXiv preprint
  arXiv:1704.04861}, 2017.

\bibitem{xie2017aggregated}
S.~Xie, R.~Girshick, P.~Doll{\'a}r, Z.~Tu, and K.~He, ``Aggregated residual
  transformations for deep neural networks,'' in \emph{Proceedings of the IEEE
  conference on computer vision and pattern recognition}, 2017, pp. 1492--1500.

\bibitem{ye2018rethinking}
J.~Ye, X.~Lu, Z.~Lin, and J.~Z. Wang, ``Rethinking the
  smaller-norm-less-informative assumption in channel pruning of convolution
  layers,'' \emph{arXiv preprint arXiv:1802.00124}, 2018.

\bibitem{molchanov2017pruning}
P.~Molchanov, S.~Tyree, T.~Karras, T.~Aila, and J.~Kautz, ``Pruning
  convolutional neural networks for resource efficient inference,'' in
  \emph{ICLR}, 2017.

\bibitem{theis2018faster}
L.~Theis, I.~Korshunova, A.~Tejani, and F.~Huszar, ``Faster gaze prediction
  with dense networks and fisher pruning,'' \emph{arXiv preprint
  arXiv:1801.05787}, 2018.

\bibitem{he2020learning}
Y.~He, Y.~Ding, P.~Liu, L.~Zhu, H.~Zhang, and Y.~Yang, ``Learning filter
  pruning criteria for deep convolutional neural networks acceleration,'' in
  \emph{CVPR}, 2020, pp. 2009--2018.

\bibitem{louizos2017learning}
C.~Louizos, M.~Welling, and D.~P. Kingma, ``Learning sparse neural networks
  through $ l\_0 $ regularization,'' in \emph{ICLR}, 2018.

\bibitem{kang2020operation}
M.~Kang and B.~Han, ``Operation-aware soft channel pruning using differentiable
  masks,'' in \emph{International Conference on Machine Learning}.\hskip 1em
  plus 0.5em minus 0.4em\relax PMLR, 2020, pp. 5122--5131.

\bibitem{herrmann2020channel}
C.~Herrmann, R.~S. Bowen, and R.~Zabih, ``Channel selection using gumbel
  softmax,'' in \emph{European Conference on Computer Vision}.\hskip 1em plus
  0.5em minus 0.4em\relax Springer, 2020, pp. 241--257.

\bibitem{huang2018data}
Z.~Huang and N.~Wang, ``Data-driven sparse structure selection for deep neural
  networks,'' in \emph{ECCV}, 2018, pp. 304--320.

\bibitem{xiao2019autoprune}
X.~Xiao, Z.~Wang, and S.~Rajasekaran, ``Autoprune: Automatic network pruning by
  regularizing auxiliary parameters,'' \emph{Advances in neural information
  processing systems}, vol.~32, 2019.

\bibitem{jang2017categorical}
E.~Jang, S.~Gu, and B.~Poole, ``Categorical reparameterization with
  gumbel-softmax,'' in \emph{ICLR}, 2017.

\bibitem{maddison2017concrete}
C.~J. Maddison, A.~Mnih, and Y.~W. Teh, ``The concrete distribution: A
  continuous relaxation of discrete random variables,'' in \emph{ICLR}, 2017.

\bibitem{bengio2013estimating}
Y.~Bengio, N.~L{\'e}onard, and A.~Courville, ``Estimating or propagating
  gradients through stochastic neurons for conditional computation,''
  \emph{arXiv preprint arXiv:1308.3432}, 2013.

\bibitem{courbariaux2016binarized}
M.~Courbariaux, I.~Hubara, D.~Soudry, R.~El-Yaniv, and Y.~Bengio, ``Binarized
  neural networks: Training deep neural networks with weights and activations
  constrained to+ 1 or-1,'' \emph{arXiv preprint arXiv:1602.02830}, 2016.

\bibitem{yin2019blended}
P.~Yin, S.~Zhang, J.~Lyu, S.~Osher, Y.~Qi, and J.~Xin, ``Blended coarse
  gradient descent for full quantization of deep neural networks,''
  \emph{Research in the Mathematical Sciences}, vol.~6, no.~1, pp. 1--23, 2019.

\bibitem{cai2017deep}
Z.~Cai, X.~He, J.~Sun, and N.~Vasconcelos, ``Deep learning with low precision
  by half-wave gaussian quantization,'' in \emph{Proceedings of the IEEE
  conference on computer vision and pattern recognition}, 2017, pp. 5918--5926.

\bibitem{yin2019understanding}
P.~Yin, J.~Lyu, S.~Zhang, S.~Osher, Y.~Qi, and J.~Xin, ``Understanding
  straight-through estimator in training activation quantized neural nets,''
  \emph{arXiv preprint arXiv:1903.05662}, 2019.

\bibitem{hinton2012neural}
G.~Hinton, ``Neural networks for machine learning coursera video lectures,''
  2012.

\bibitem{he2015delving}
K.~He, X.~Zhang, S.~Ren, and J.~Sun, ``Delving deep into rectifiers: Surpassing
  human-level performance on imagenet classification,'' in \emph{Proceedings of
  the IEEE international conference on computer vision}, 2015, pp. 1026--1034.

\bibitem{tan2019efficientnet}
M.~Tan and Q.~V. Le, ``Efficientnet: Rethinking model scaling for convolutional
  neural networks,'' in \emph{ICLR}, 2019.

\bibitem{yannMNIST}
Y.~LeCun, C.~Corinna, and B.~Christopher~J.C., ``The mnist database of
  handwritten digits,'' \url{http://yann.lecun.com/exdb/mnist/index.html}.

\bibitem{krizhevsky2009learning}
A.~Krizhevsky and G.~Hinton, ``Learning multiple layers of features from tiny
  images,'' University of Toronto, Tech. Rep., 2009.

\bibitem{ILSVRC15}
O.~Russakovsky, J.~Deng, H.~Su, J.~Krause, S.~Satheesh, S.~Ma, Z.~Huang,
  A.~Karpathy, A.~Khosla, M.~Bernstein, A.~C. Berg, and L.~Fei-Fei, ``{ImageNet
  Large Scale Visual Recognition Challenge},'' \emph{International Journal of
  Computer Vision}, vol. 115, no.~3, pp. 211--252, 2015.

\bibitem{merity2017pointer}
S.~Merity, C.~Xiong, J.~Bradbury, and R.~Socher, ``Pointer sentinel mixture
  models,'' in \emph{ICLR}, 2017.

\bibitem{liu2019rethinking}
Z.~Liu, M.~Sun, T.~Zhou, G.~Huang, and T.~Darrell, ``Rethinking the value of
  network pruning,'' \emph{ICLR}, 2019.

\bibitem{inan2017tying}
H.~Inan, K.~Khosravi, and R.~Socher, ``Tying word vectors and word classifiers:
  A loss framework for language modeling,'' in \emph{ICLR}, 2017.

\bibitem{ioffe2015batch}
S.~Ioffe and C.~Szegedy, ``Batch normalization: Accelerating deep network
  training by reducing internal covariate shift,'' \emph{arXiv preprint
  arXiv:1502.03167}, 2015.

\bibitem{van2017l2}
T.~van Laarhoven, ``L2 regularization versus batch and weight normalization,''
  in \emph{Advances in neural information processing systems}, 2017.

\bibitem{yu2018nisp}
R.~Yu, A.~Li, C.-F. Chen, J.-H. Lai, V.~I. Morariu, X.~Han, M.~Gao, C.-Y. Lin,
  and L.~S. Davis, ``Nisp: Pruning networks using neuron importance score
  propagation,'' in \emph{Proceedings of the IEEE Conference on Computer Vision
  and Pattern Recognition}, 2018, pp. 9194--9203.

\bibitem{zagoruyko2016wide}
S.~Zagoruyko and N.~Komodakis, ``Wide residual networks,'' in \emph{BMVC},
  2016.

\bibitem{hu2016network}
H.~Hu, R.~Peng, Y.-W. Tai, and C.-K. Tang, ``Network trimming: A data-driven
  neuron pruning approach towards efficient deep architectures,'' \emph{arXiv
  preprint arXiv:1607.03250}, 2016.

\bibitem{narang2017exploring}
S.~Narang, G.~Diamos, S.~Sengupta, and E.~Elsen, ``Exploring sparsity in
  recurrent neural networks,'' in \emph{ICLR}, 2017.

\bibitem{li2018independently}
S.~Li, W.~Li, C.~Cook, C.~Zhu, and Y.~Gao, ``Independently recurrent neural
  network (indrnn): Building a longer and deeper rnn,'' in \emph{CVPR}, 2018,
  pp. 5457--5466.

\end{thebibliography}
%



%




\vspace{-10mm}
\begin{IEEEbiography}
[{\includegraphics[width=1in,height=1.25in,clip,keepaspectratio]{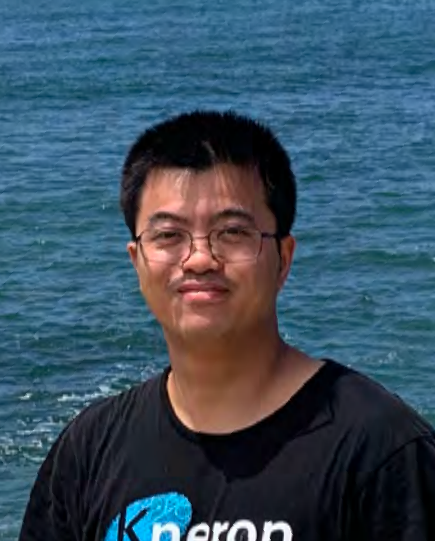}}]
{Zhimin Tang} received B.S. degree from Hunan Normal University in 2012 and M.S. degree from Hunan University in 2015. He is currently pursuing the Ph.D. degree in the Department of Automation at Xiamen University, China. He was a visiting scholar at the Department of Electrical and Computer Engineering of Southern Illinois University Carbondale from 2018 to 2020. His research interests include image processing, machine learning, and computer vision.
\end{IEEEbiography}
\begin{IEEEbiography}
 [{\includegraphics[width=1in,height=1.25in,clip,keepaspectratio]{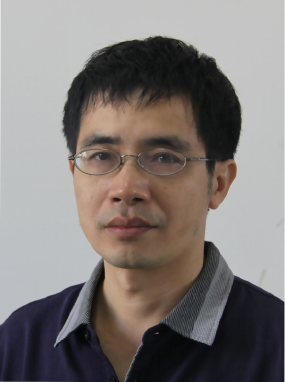}}]
{Linkai Luo} is a Professor in the Department of Automation, Xiamen University, China. His research interests include machine learning, pattern recognition, data mining, computer vision, biological information processing and financial data analysis.
\end{IEEEbiography}
\begin{IEEEbiography}[{\includegraphics[width=1in,height=1.25in,clip,keepaspectratio]{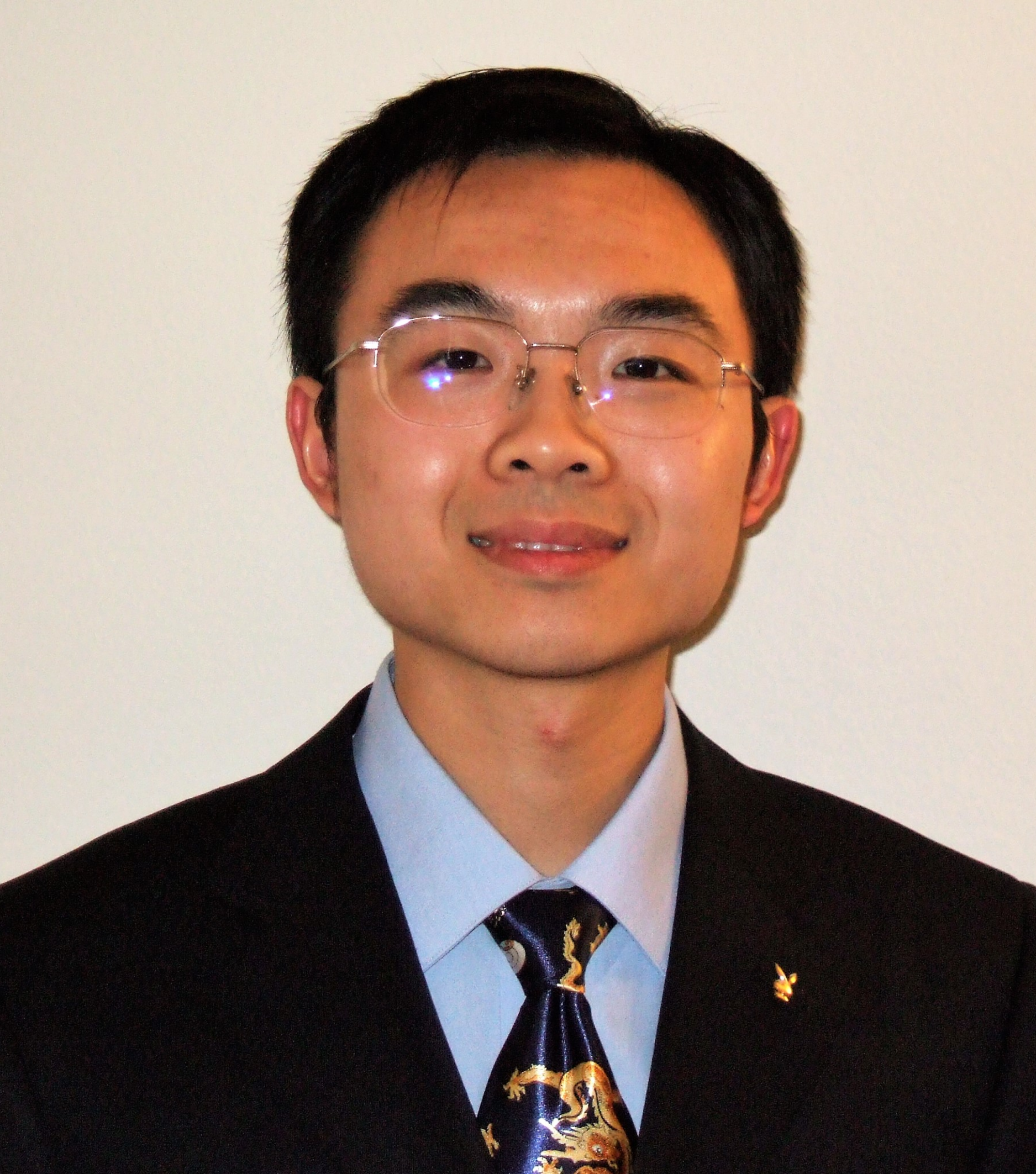}}]{Bike Xie}
is currently a director of engineering in Kneron Inc, San Diego. He is and has been working on system architecture design for AI chip, deep learning model compression, and algorithm design for deep learning applications. He received his B.S. degree in electronic engineering from Tsinghua University, Beijing, in 2005. He received his M.S. and the Ph.D. degrees in electrical engineering from the University of California, Los Angeles in 2006 and 2010 respectively. At UCLA, he worked on a broad range of research topics including capacity regions and encoding schemes for broadcast channels, download-time regions for peer-to-peer networks, packet coding and exchange for broadcast networks, universal turbo codes for space-time channels, and channel code design for optical communications. Dr. Xie then joined Marvell Semiconductor Inc., Santa Clara in 2010. At Marvell, he led a system team design physical layer systems and specifications for 6Gbps to 112Gbps SerDes systems, and capacitive touchscreen systems. In 2017, he joined Kneron Inc.
\end{IEEEbiography}
\vspace{-10mm}
\begin{IEEEbiography}[{\includegraphics[width=1in,height=1.25in,clip,keepaspectratio]{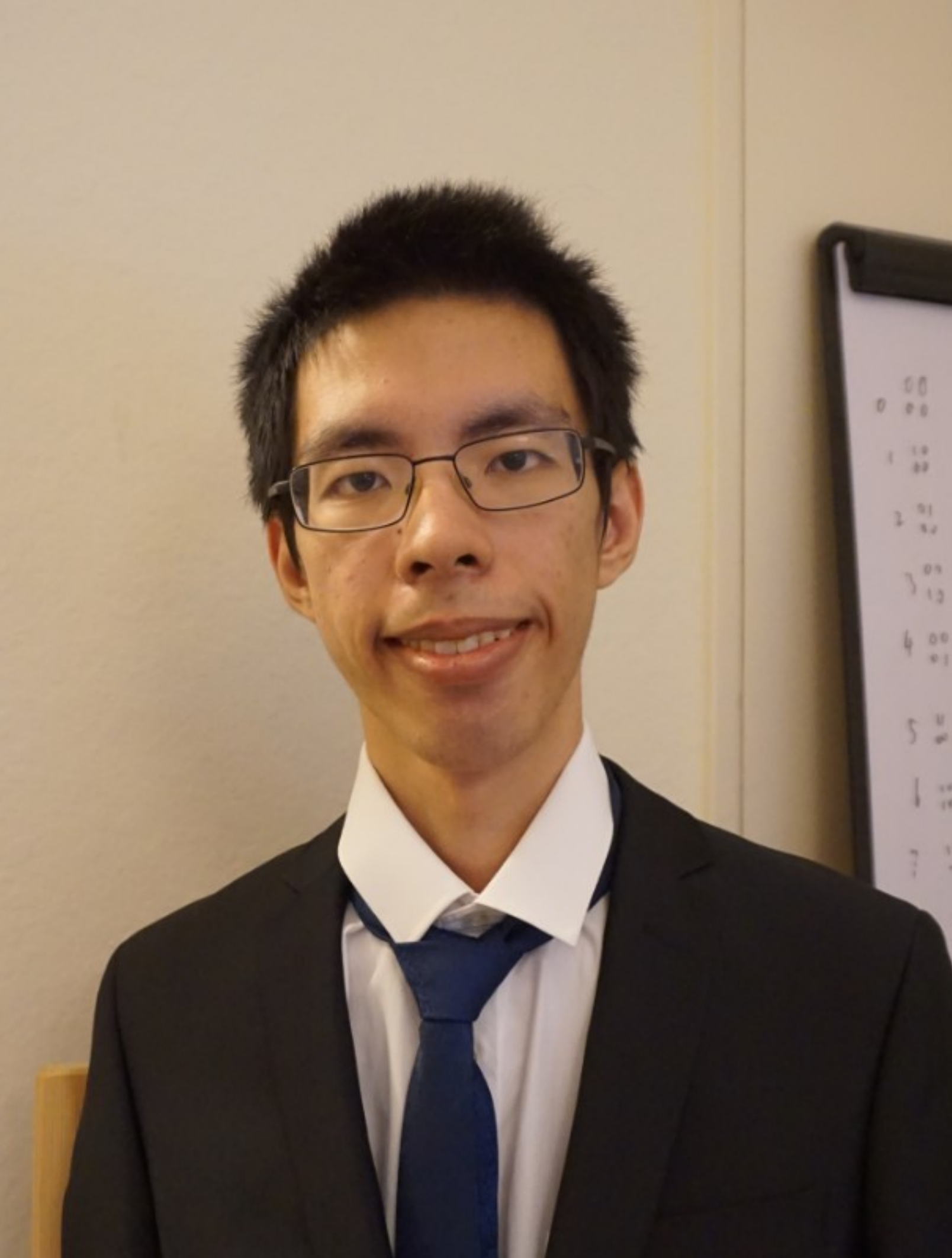}}]{Yiyu Zhu}
received B.S. degree and M.S degree in Electrical Engineering from University of California, San Diego in 2017 and 2018. He worked at Kneron Inc as an algorithm engineer between 2018 and 2021. During this time, he developed algorithm for model compression and inference acceleration. His research interests include machine learning, computer architecture, and high performance computing.
\end{IEEEbiography}
\vspace{-10mm}
\begin{IEEEbiography}[{\includegraphics[width=1in,height=1.25in,clip,keepaspectratio]{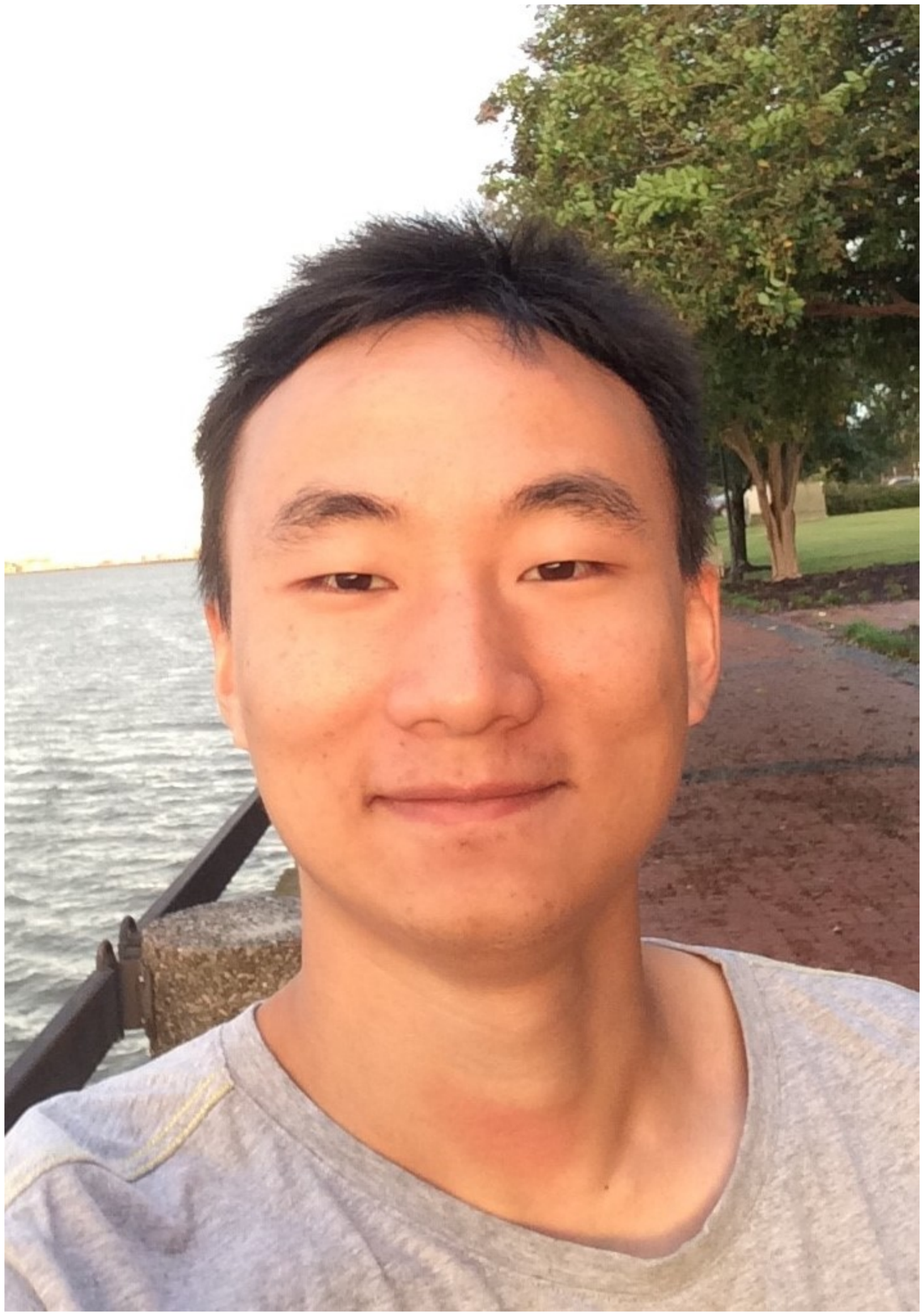}}]{Rujie Zhao}
received a B.S. degree in Electrical Engineering from Jiang Su Normal University and an M.S. degree from the State University of New York at New Paltz. Since fall 2018, he is working on his Ph.D. degree in the Department of Electrical and Computer Engineering at Southern Illinois University Carbondale, Carbondale, IL, USA. His research interests include deep neural network algorithms and architectures, memristor-based neural network implementation and optimization.
\end{IEEEbiography}
\vspace{-10mm}
\begin{IEEEbiography}[{\includegraphics[width=1in,height=1.25in,clip,keepaspectratio]{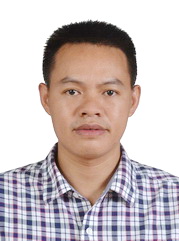}}]{Lvqing Bi}
received the B.S., M.S. degrees in Electronic Science and technology from the Guangxi University, Nanning, China, in 2003 and 2013, respectively. Since 2015, he has been working toward the Ph.D. degree in Electronic Science and technology at the School of Electronic Science and Engineering, Xiamen University, Xiamen, China. His current research interests include artificial neural networks, intelligent manufacturing technology, network signal processing, terahertz metamaterials, information entropy and fuzzy sets.
\end{IEEEbiography}
\vspace{-10mm}
\begin{IEEEbiography}
[{\includegraphics[width=1in,height=1.25in,clip,keepaspectratio]{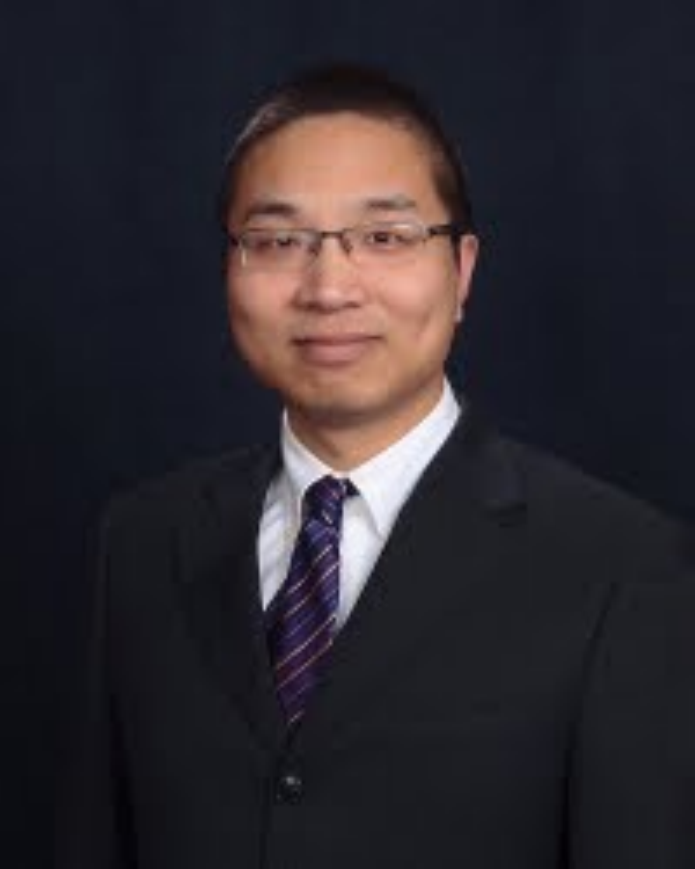}}]
{Chao Lu} received B.S. degree in electrical engineering from Nankai University and M.S. degree from the Hong Kong University of Science and Technology. He obtained Ph.D. degree at Purdue University, West Lafayette in 2012. From 2013 to 2015, he worked in US industry. Since July 2020, he became an associate professor of the Electrical and Computer Engineering Department of Southern Illinois University, Carbondale, IL, USA. His research interests include the design of high-performance circuit and system design, and deep machine learning algorithms. Mr. Lu was the recipient of the Best Paper Award of the International Symposium on Low Power Electronics and Design in 2007, Best Paper Award Nomination of IEEE System-on-Chip Conference (2016), and Top 5 team in the International Hardware Design Contest of IEEE Design Automation Conference (2017).
\end{IEEEbiography}



\end{document}